\pdfoutput=1
\documentclass[11pt]{article}

\usepackage[preprint]{acl}

\usepackage{times}
\usepackage{latexsym}
\usepackage[T1]{fontenc}
\usepackage[utf8]{inputenc}
\usepackage{microtype}

\usepackage{booktabs}       %
\usepackage{colortbl}       %
\usepackage{multirow}       %
\usepackage{array}          %
\usepackage{float}          %
\usepackage{amsmath}        %
\usepackage{amssymb}
\usepackage{graphicx}
\usepackage{mdframed}       %
\usepackage{listings}       %
\usepackage{enumitem}       %
\usepackage{tcolorbox}      %
\usepackage{orcidlink}      %
\tcbuselibrary{listings,skins,breakable}
\usepackage{pgfplots}       %
\pgfplotsset{compat=1.18}

\definecolor{hdrbg}{HTML}{1A2940}
\definecolor{bbfill}{HTML}{D6EAF8}
\definecolor{wbfill}{HTML}{FEF9E7}
\definecolor{jdfill}{HTML}{D5F5E3}
\definecolor{sysred}{HTML}{C0392B}
\definecolor{volorange}{HTML}{E67E22}
\definecolor{hedblue}{HTML}{2980B9}
\definecolor{stocgreen}{HTML}{27AE60}
\definecolor{trivhdr}{HTML}{1F3D6A}
\definecolor{haluhdr}{HTML}{1A5C3A}
\definecolor{selfhdr}{HTML}{7B241C}
\definecolor{popqahdr}{HTML}{4A235A}
\definecolor{phcolor}{HTML}{C0392B}

\newcolumntype{L}[1]{>{\raggedright\arraybackslash}p{#1}}
\newcolumntype{C}[1]{>{\centering\arraybackslash}p{#1}}

\definecolor{codebg}{HTML}{F8F8F8}
\definecolor{codeframe}{HTML}{CCCCCC}
\definecolor{codecomment}{HTML}{6A9153}
\definecolor{codestring}{HTML}{CE9178}
\definecolor{codekw}{HTML}{0000CC}

\lstdefinestyle{pysnippet}{
  language=Python,
  basicstyle=\ttfamily\footnotesize,
  keywordstyle=\color{codekw}\bfseries,
  stringstyle=\color{codestring},
  commentstyle=\color{codecomment}\itshape,
  numberstyle=\tiny\color{gray},
  numbers=left,
  numbersep=6pt,
  stepnumber=1,
  breaklines=true,
  breakatwhitespace=false,
  showstringspaces=false,
  tabsize=4,
  frame=none,
  xleftmargin=2pt,
  xrightmargin=2pt,
}

\tcbset{
  codebox/.style={
    enhanced,
    colback=codebg,
    colframe=codeframe,
    boxrule=0.6pt,
    arc=3pt,
    top=4pt, bottom=4pt, left=4pt, right=4pt,
    breakable,
    fonttitle=\small\bfseries,
    attach boxed title to top left={xshift=8pt, yshift=-6pt},
    boxed title style={colback=hdrbg, colframe=hdrbg, arc=2pt,
                        fontupper=\small\bfseries\color{white}},
  }
}

\providecommand{\twoline}[2]{\begin{tabular}[c]{@{}c@{}}#1 \\[-1pt] {\scriptsize $\pm$#2}\end{tabular}}
\providecommand{\onepm}[2]{#1\,{\scriptsize$\pm$#2}}

\title{DECK: A Consistency × Confidence Taxonomy of LLM Hallucinations}

\author{Mohit Singh Chauhan~\orcidlink{0000-0002-7817-0427} \\
Independent Researcher \\
\texttt{mohitcek@gmail.com}}

\begin{document}
\maketitle

\begin{abstract}
Existing hallucination taxonomies classify LLM errors by \emph{what
is wrong with the output} --- memorised misconceptions, reasoning
failures, fluent fabrications --- but cannot answer a different
question: \emph{which uncertainty scorer would have caught this
error?} We propose a complementary taxonomy that classifies errors by
their \textbf{detectability signature}, the signal a scorer family
would read. The \textbf{DECK taxonomy} is a $2{\times}2$ partition
along inter-sample consistency and token-level confidence into four
regimes (\textbf{D}rift, \textbf{E}ntrenched, \textbf{C}onfabulation,
\textbf{K}notted) that yields a falsifiable blind-spot map: black-box
consistency scorers have signal in D and C, white-box
token-probability scorers in K and C, and only an LLM-as-a-Judge with
independent pretraining can detect E. Across three models and four
short-form QA datasets we test this map two ways: judge-involving
scorer disagreements concentrate in each family's predicted blind-spot
cells, and external labels (SelfAware unanswerable, HaluEval
adversarial, PopQA entity popularity) land in the predicted cells,
robustly to cross-fitted thresholds. We further identify
a \emph{universal blind spot of output-level UQ}: on knowledge-gap
inputs where the generator emits confident, repeatable fabrications,
every output-level family collapses by construction. A linear probe on
Llama-3-8B's final-layer hidden states also falls to chance, with or
without quantisation, though an intermediate layer retains weak signal.

\end{abstract}

\section{Introduction}
\label{sec:intro}
Existing hallucination taxonomies classify LLM errors by
\emph{what is wrong with the
output}~\citep{ji2023survey,huang2025hallucination,wang2024factuality}
--- memorised misconceptions, reasoning failures, fluent
fabrications, faithfulness violations. These taxonomies are useful
for diagnosis, but they cannot answer a question that arises whenever
an LLM is deployed in a high-stakes setting: \emph{which uncertainty
scorer would have caught this error?} The same content category can
produce different detection signatures across models (a ``reasoning
failure'' is detectable by a Judge if the answer is stable and
high-probability, but only by an inter-sample consistency check if
the answer varies); and the same detection signature can come from
very different content categories. Without a taxonomy on the
detection axis, choosing a scorer for a new domain stays a guess ---
even though the literature already offers three scorer families:
\textbf{Black-box} (BB) consistency
checks~\citep{manakul2023selfcheckgpt,kuhn2023semantic},
\textbf{White-box} (WB) token-level
log-probabilities~\citep{kadavath2022language,malinin2020uncertainty},
and \textbf{LLM-as-a-Judge} (J) factual
review~\citep{zheng2023judging}, which have been shown to ensemble
well together~\citep{bouchard2025uqlm}.

We propose a complementary taxonomy that classifies errors by their
detectability signature --- the signal a scorer family
reads. Each of our two axes already underpins existing scorers
(consistency for BB, token probability for WB); the contribution is
what follows from crossing them explicitly: a \emph{falsifiable
per-family blind-spot map} (\S\ref{sec:taxonomy}) that predicts which
family will miss which hallucination, and two independent tests of
those predictions (\S\ref{sec:disagreement}, \S\ref{sec:externalsignal})
that could have failed but did not.

\textbf{(1) The DECK taxonomy} (Table~\ref{tab:taxonomy}): a
$2{\times}2$ partition along inter-sample consistency and token-level
confidence into four behavioural regimes (D,
E, C, K).
Each cell maps to a specific scorer family (or families) that can detect
it: BB has signal in D and C; WB has signal in K and C; only J can
detect E (and only when the judge has independent pretraining). For
each scorer axis we pick the threshold that best separates correct
from hallucinated responses (Youden's J), then assign each sample to
one of the four cells. We test the map in two ways: judge-involving
scorer disagreements --- the Judge being a third axis not used to
define the cells --- should concentrate in each family's predicted
blind-spot cells (\S\ref{sec:disagreement}); and external labels
(SelfAware unanswerable, HaluEval adversarial, PopQA entity
popularity), which involve no scorer beyond the cell split, should
land in the predicted cells (\S\ref{sec:externalsignal}). Both hold,
with model-scale and content-specific secondary-cell refinements, and
the cell assignment is stable under cross-fitted thresholds
(Appendix~\ref{app:threshold}). Our experiments cover short-form QA;
\S\ref{sec:longform} outlines the extension to long-form generation.
\textbf{(2) A universal blind spot of output-level UQ.} We identify
an empirical regime where every scorer in the three-family
output-level paradigm fails simultaneously
(\S\ref{sec:universalfailure}). On SelfAware all three families
collapse to or below chance because the generator emits confident,
repeatable fabrications on inputs it cannot answer --- eliminating
the variance every output-level family needs to be informative. The
right engineering response is an abstention envelope that routes such
out-of-scope inputs to refusal before scoring. ``Universal'' here means across the three output-level families studied.
We also ask whether the failure persists in the activations: a linear
probe on Llama-3-8B's last-layer hidden states falls to chance on
SelfAware with or without 4-bit quantisation, but an intermediate layer
retains weak signal (\S\ref{sec:universalfailure}), which bounds the
blind spot to output-level signals.
\begin{table*}[ht]
\centering
\caption{The \textbf{DECK taxonomy} of LLM hallucinations. Each cell is a
behavioral regime defined by two observable axes: inter-sample consistency
and token-level confidence. The four regimes map to the letters
\textbf{D}rift, \textbf{E}ntrenched, \textbf{C}onfabulation, and
\textbf{K}notted. The final column gives the scorer family (or families)
sensitive to each regime.
}
\label{tab:taxonomy}
\setlength{\tabcolsep}{6pt}
\renewcommand{\arraystretch}{1.45}
\scriptsize
\begin{tabular}{%
  >{\bfseries\centering\arraybackslash}p{2.0cm}%
  L{4.2cm}%
  L{4.2cm}%
  L{3.9cm}%
}
\rowcolor{hdrbg}
\textcolor{white}{\textbf{}} &
\textcolor{white}{\textbf{Low Consistency}} &
\textcolor{white}{\textbf{High Consistency}} &
\textcolor{white}{\textbf{Detectable By}} \\[1pt]
\cellcolor{bbfill}High Token\newline Confidence &
\cellcolor{white}\textcolor{volorange}{\textbf{D --- Drift}}\newline
    \emph{Different confident wrong answer each sample.} Multiple
    plausible-but-wrong answers are drawn confidently from the output
    distribution; each draw is internally coherent, but the answer drifts. &
\cellcolor{white}\textcolor{sysred}{\textbf{E --- Entrenched}}\newline
    \emph{Same confident wrong answer every sample.} Model has locked
    onto a memorized misconception or shared-pretraining error and
    reproduces it without variance. &
\cellcolor{jdfill}\textbf{Drift:} Black-box and Judge. \par\textbf{Entrenched:} Judge (independent provider required); blind to both BB and WB. \\[3pt]
\hline
\cellcolor{bbfill}Low Token\newline Confidence &
\cellcolor{white}\textcolor{stocgreen}{\textbf{C --- Confabulation}}\newline
    \emph{Different low-probability wrong answer each sample.} Classic
    uncertainty: the
    model genuinely does not know, and different samples produce
    different wrong answers with low confidence. &
\cellcolor{white}\textcolor{hedblue}{\textbf{K --- Knotted}}\newline
    \emph{Same low-probability wrong answer every sample.} Model is
    consistently unsure: it settles on the same hedged answer each time
    but assigns low token probability. &
\cellcolor{jdfill}
    \textbf{Confabulation:} All three families.\par
    \textbf{Knotted:} White-box and Judge.  \\
\hline
\end{tabular}
\vspace{2pt}
\begin{minipage}{0.95\linewidth}
\footnotesize\emph{Note.} Regimes are defined by observable output
behavior, not by an internal mechanism claim. Cell membership depends on
the scorer thresholds used to discretize the two axes; described in \S\ref{sec:taxonomy}.
\end{minipage}
\end{table*}

\section{Background and Related Work}
\label{sec:related}
\subsection{Three Families of UQ Scorers}
\label{sec:bg-families}

Output-level UQ methods for LLMs fall into three families distinguished by the
access they require~\citep{shorinwa2025survey, kang2025uqsurvey}.
\emph{White-box} methods read confidence from token log-probabilities;
representative scorers include \texttt{P(True)}~\citep{kadavath2022language},
sequence-probability measures~\citep{malinin2020uncertainty}, and
calibration-aware variants over token
distributions~\citep{jiang2021know, duan2024shifting}. They require
logit access and are cheap. A related thread elicits \emph{verbalized}
confidence by asking the model to state its own
uncertainty~\citep{tian2023askforcalibration, xiong2024llms}.
\emph{Black-box} methods read confidence from inter-sample agreement:
SelfCheckGPT~\citep{manakul2023selfcheckgpt} uses NLI- and embedding-based
consistency, while semantic entropy~\citep{kuhn2023semantic,
farquhar2024detecting} clusters responses into meaning classes and computes
entropy over the resulting distribution. Subsequent work explores
alternative aggregations of the consistency signal: pairwise-NLI
aggregation across sampled responses~\citep{lin2023generating} and
semantic-density estimation in embedding
space~\citep{qiu2024semantic}. These methods are model-agnostic but
cost multiple generations per query. \emph{LLM-as-a-Judge} methods delegate evaluation to a separate
model~\citep{zheng2023judging, liu2023geval}; they detect failures invisible to
either probabilistic signal but inherit known judge biases~\citep{chiang2023can}
and risk shared-pretraining errors when generator and judge are
related~\citep{feng2024dont}. Recent open-source aggregators
that combine all three families have shown empirically that
ensembles outperform individual scorers but offer no mechanistic
account of \emph{why} a given family complements a Judge. A fourth,
internal-state family --- UQ heads~\citep{shelmanov2025uqhead},
hidden-state probes~\citep{slobodkin2023curious}, and
information-theoretic estimators~\citep{yadkori2024believe} --- reads
the residual-stream activations that produce the token logits a
white-box scorer would consume. We treat internal-state UQ as a
deeper access tier of the same signal flow rather than a fundamentally
distinct family; one representative is evaluated as a robustness
check in \S\ref{sec:universalfailure}, broader internal-state methods
are out of scope.

\subsection{Aleatoric, Epistemic, and Confabulation}
\label{sec:bg-aleaepi}

The classical aleatoric/epistemic
decomposition~\citep{hullermeier2021aleatoric} remains useful at LLM scale:
epistemic uncertainty is the component most tightly linked to
hallucination~\citep{yadkori2024believe, shorinwa2025survey}.
Across black-box UQ work, the canonical failure mode of interest is
\emph{confabulation}: fluent, wrong outputs that vary across
stochastic samples (typically obtained by temperature sampling rather
than re-seeded decoding). \citet{farquhar2024detecting} formalise
this notion and validate semantic entropy against it, but
consistency-based scoring across SelfCheckGPT, semantic entropy, and
their successors all target the same underlying error class. We
adopt the term \emph{confabulation} unchanged for our
low-consistency, low-confidence cell; the categories Farquhar et al.\
set aside (misconceptions, lies~\citep{evans2021truthful}, reasoning
failures) share a high-consistency, high-confidence signature and
constitute our \emph{Entrenched} cell. Our contribution is to make this implicit partition
explicit as a full $2{\times}2$ taxonomy and use it to explain the
complementarity patterns that prior ensembling work reports but does
not analyse mechanistically.

\section{A Taxonomy of LLM Hallucination Types}
\label{sec:taxonomy}
\subsection{Two-Axis Framework}

We characterize LLM hallucinations along two observationally independent axes:
inter-sample consistency (does the model produce the same incorrect answer across
multiple independent samples?) and token-level confidence (does the model assign
high probability to the tokens of its generated response?). This yields four hallucination
types, summarized in Table~\ref{tab:taxonomy}, each implying a distinct detection strategy.

\subsection{Why Each Family Has a Blind Spot}

BB scorers measure consistency: if a model produces the same wrong answer
every sample, the scorer sees high agreement and assigns high confidence---\emph{Entrenched
and Knotted hallucinations evade detection.}
The Knotted blind spot is sharper than it might appear: high
inter-sample agreement looks like high confidence to a BB scorer
\emph{regardless} of token-level probability, so even a low-probability
``locked-in'' hedge that the model repeats every sample (the Knotted
signature) registers as confident-and-correct. WB scorers measure token
probability: a model may assign high probability to a fluent but factually incorrect
claim---\emph{Entrenched and Drift hallucinations evade detection.} LLM judges
evaluate factual content and in principle detect any type when the judge's knowledge is
accurate. However, they are susceptible to the same systematic misconceptions as the generator
when both share pretraining data---\emph{shared Entrenched hallucinations are the failure
mode.} More strikingly, on knowledge-gap inputs (questions genuinely unanswerable
for any LLM) on which the generator nevertheless emits a confident,
repeatable fabrication, all three families fail \emph{by construction}:
BB scorers see consistent confident answers, WB scorers see high token
probabilities, and judges share the same knowledge gap. This represents
a regime --- the \emph{universal blind spot} of output-level
UQ --- that generally evades detection by any output-level scorer family.
This asymmetric coverage motivates the per-pair disagreement analysis
in \S\ref{sec:disagreement}, which tests whether the predicted
quadrant signatures emerge empirically.

\subsection{Extending DECK to Long-Form Generation}
\label{sec:longform}

Our experiments use short-form QA, where one response carries one
factual claim. In long-form generation a response bundles many claims
that can sit in different cells. DECK extends naturally by moving
both axes from the response to the claim. Following
FActScore~\citep{min2023factscore}, factor each response into atomic
claims; measure \emph{claim-level consistency} as the fraction of
stochastic samples that contain a matching or entailed claim
\citep{manakul2023selfcheckgpt,zhang-etal-2024-luq}; and measure
\emph{claim-level confidence} from the token probabilities of the
span that expresses the claim. Each claim then receives its own cell,
and a response becomes a \emph{cell mixture}: a biography might carry an
Entrenched claim that repeats a common misconception alongside
Confabulation claims about obscure dates. The blind-spot predictions carry over per claim,
so the mixture indicates which scorer families a given response
needs. Two problems remain open: \emph{claim alignment} across
stochastic samples that paraphrase or reorder content, and
\emph{confidence attribution} when the same claim is realised by
different token spans. We leave claim-level validation to future
work; all headline claims in this paper are scoped to short-form QA.

\section{Experimental Setup}
\label{sec:setup}
\subsection{Datasets}

\textbf{TriviaQA} \citep{joshi2017triviaqa} contains factual trivia where larger
models are prone to memorizing systematic misconceptions---expected to exercise
the Entrenched cell. \textbf{HaluEval} \citep{li2023halueval} provides curated
hallucinated answer pairs with binary labels, skewing toward Entrenched and
Knotted types. \textbf{SelfAware} \citep{yin2023selfaware}; we use a 500-question subset of its unanswerable
questions to probe model abstention behavior, targeting the
Entrenched cell for knowledge-gap hallucinations. \textbf{PopQA}
\citep{mallen2023popqa} contains entity-centric questions stratified by
Wikipedia page-view popularity; the popular- and rare-entity slices target the
Entrenched and Confabulation cells, respectively.
Response grading procedures for all four datasets are detailed in
Appendix~\ref{app:grading}.

\subsection{Models}

We evaluate three models spanning a range of scale and provider:
\textbf{Llama-3-8B} (smaller, open-weights),
\textbf{GPT-4o} (larger, closed-weights), and
\textbf{Gemini-2.5-Flash} (mid-scale, closed-weights, providing a third provider to test
generality). All models are evaluated on all four datasets: TriviaQA, HaluEval, SelfAware,
and PopQA.

\subsection{Scorer Selection}

We compute confidence scores using the UQLM toolkit \citep{bouchard2026uqlm} across 6 black-box, 5 white-box, and 4 judge configurations, giving 15 scorers per model-dataset pair
(full AUROC breakdown in Appendix~\ref{app:auroc}, Table~\ref{tab:auroc5}). For the
disagreement and complementarity analysis, each family is represented by its single
highest-AUROC method on the target split, avoiding hand-selection bias. A full
reference table is provided in Appendix~\ref{app:scorers}.

\medskip\noindent
\textbf{Black-box:} Semantic Negentropy, Non-Contradiction, Entailment,
Cosine Similarity, Exact Match, Semantic Sets Confidence.

\medskip\noindent
\textbf{White-box:} Sequence Probability, Min Token Probability, Mean Token
Negentropy, Probability Margin, P(True).

\medskip\noindent
\textbf{LLM-as-a-Judge:} Four judges per generator span increasing
generator independence: the model itself (self-judge, lower
bound on utility), the other two generators (cross-LLM
judges, reducing shared-pretraining bias), and Claude Sonnet~4.6
(independent-provider, upper bound on independence).
All judges use a Likert scoring template; multi-generation scorers
use five sampled responses at temperature~1.0; $k\!=\!15$ for top-$k$
token-probability scorers.

\subsection{Evaluation Metrics}
 
We measure detector quality with AUROC per scorer (Appendix~\ref{app:auroc},
Table~\ref{tab:auroc5}) and complementarity between scorer families with
the hallucination-restricted complementarity score
$C_H(A,B)$. Let $\mathcal{H} = \{i : \neg\,\text{correct}_i\}$ be the
hallucinated-sample subset of size $N_H = |\mathcal{H}|$.
$C_H(A,B)$ is the fraction of $\mathcal{H}$ on which the binary
predictions of scorer $A$ and $B$ disagree:
\begin{align}
  C_H(A,B) = \frac{|\{i \in \mathcal{H} :
      \hat{y}_i^{(A)} \neq \hat{y}_i^{(B)}\}|}{N_H}
  \label{eq:comp_hal}
\end{align}
where $\hat{y}_i^{(A)}=\mathbf{1}[s_i^{(A)}<\tau^{(A)}]$ is scorer
$A$'s binary hallucination prediction at threshold $\tau^{(A)}$. We
choose $\tau^{(A)}$ as the Youden's J optimal threshold
$\tau^{(A)}=\arg\max_\tau\,[\mathrm{TPR}(\tau)-\mathrm{FPR}(\tau)]$
fit on the full $N\!=\!500$ samples of each condition (positive class
$=$ hallucinated). YJ is the standard ROC-corner criterion: it
maximises balanced accuracy and is base-rate robust, so it does not
collapse on imbalanced datasets the way (e.g.) F1-optimal thresholds
do. The same threshold defines DECK cell membership
(\S\ref{sec:taxonomy}); a step-by-step computation example is provided
in Appendix~\ref{app:complementarity}. Cell assignments are stable under
5-fold cross-fitted YJ (per-sample $\kappa\!=\!0.70$--$0.96$; dominant
cell unchanged in $11/12$ external-signal rows); median and F1-optimal
alternatives are compared in Appendix~\ref{app:threshold}.

\section{Results}
\label{sec:results}

\subsection{Individual Scorer Performance}
\begin{figure*}[t]
\centering
\includegraphics[width=\textwidth]{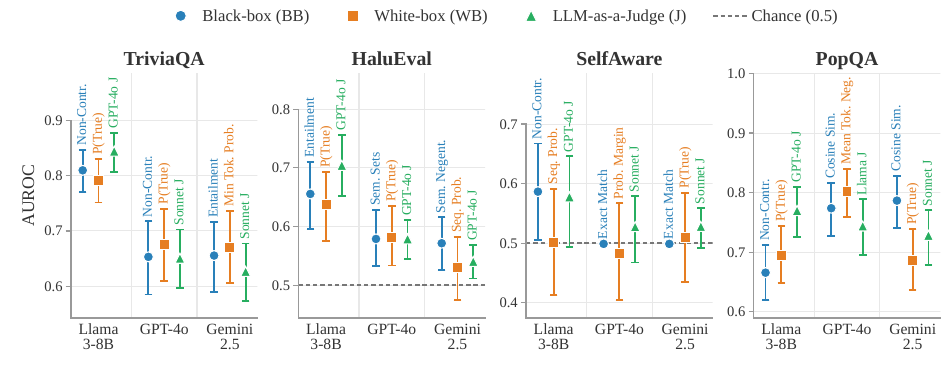}
\caption{Best-AUROC scorer per family for all twelve dataset--model
combinations, with 95\% bootstrap CIs. Marker shape encodes family
(\textbullet\,BB, $\blacksquare$\,WB, $\blacktriangle$\,Judge); the
selected scorer is labelled above each marker. Full per-method bar plots in
Figure~\ref{fig:auroc_all} and exact values with CIs in
Table~\ref{tab:auroc5} (Appendix~\ref{app:auroc}).}
\label{fig:auroc_best}
\end{figure*}

Figure~\ref{fig:auroc_best} shows the top-AUROC scorer per family across
all twelve dataset--model combinations; per-method numerical values with
$95\%$ bootstrap CI are in Table~\ref{tab:auroc5}
(Appendix~\ref{app:auroc}).
Three patterns stand out.

First, no scorer family Pareto-dominates: on Llama-3-8B TriviaQA the Judge leads
(0.844), while on GPT-4o TriviaQA WB leads (P(True) 0.675); on PopQA the BB Cosine
Similarity reaches 0.77--0.79 for GPT-4o and Gemini-2.5-Flash (and WB Mean Token
Negentropy 0.80 for GPT-4o), but the Judge takes the lead on Llama-3-8B (GPT-4o cross-judge 0.770).
Second, absolute AUROC compresses with model scale: top scorers from families that span 0.79--0.84 on
Llama-3-8B TriviaQA cluster at 0.65--0.68 on GPT-4o and 0.63--0.67 on Gemini-2.5-Flash,
consistent with the Entrenched-shift predicted by the DECK taxonomy.
Third, PopQA has the highest per-condition top-scorer AUROC on average and SelfAware the lowest; on SelfAware all
methods collapse to near-random across all three models (detailed in
\S\ref{sec:universalfailure}).
Since no family dominates, ensemble gains depend on structural disagreement, not simply on
selecting the highest-AUROC scorer---motivating \S\ref{sec:disagreement}.

\begin{table}[t]
\centering
\caption{Top-AUROC scorer in each family (BB / WB / Judge) for every
(model, dataset) combination.}
\label{tab:topscorers}
\renewcommand{\arraystretch}{1.2}
\setlength{\tabcolsep}{1pt}
\resizebox{\columnwidth}{!}{%
\tiny
\begin{tabular}{@{}C{0.75cm} L{1.3cm} C{2.0cm} C{2.0cm} C{2.0cm}@{}}
\toprule
\rowcolor{hdrbg}
\textcolor{white}{\textbf{Dataset}} &
\textcolor{white}{\textbf{Model}} &
\textcolor{white}{\textbf{BB}} &
\textcolor{white}{\textbf{WB}} &
\textcolor{white}{\textbf{Judge}} \\
\midrule
\multirow{3}{*}{\rotatebox[origin=c]{90}{TriviaQA}}
  & \cellcolor{bbfill}Llama-3-8B & \cellcolor{bbfill}Non-Cont.\ (0.810) & \cellcolor{bbfill}P(True) (0.791) & \cellcolor{bbfill}GPT-4o (0.844) \\
  & \cellcolor{wbfill}GPT-4o     & \cellcolor{wbfill}Non-Cont.\ (0.653) & \cellcolor{wbfill}P(True) (0.675) & \cellcolor{wbfill}Sonnet (0.651) \\
  & \cellcolor{jdfill}Gemini-2.5 & \cellcolor{jdfill}Entailment (0.655) & \cellcolor{jdfill}Min Prob.\ (0.670) & \cellcolor{jdfill}Sonnet (0.627) \\
\midrule
\multirow{3}{*}{\rotatebox[origin=c]{90}{HaluEval}}
  & \cellcolor{bbfill}Llama-3-8B & \cellcolor{bbfill}Entailment (0.656) & \cellcolor{bbfill}P(True) (0.638) & \cellcolor{bbfill}GPT-4o (0.704) \\
  & \cellcolor{wbfill}GPT-4o     & \cellcolor{wbfill}Sem.\ Sets (0.579) & \cellcolor{wbfill}P(True) (0.582) & \cellcolor{wbfill}GPT-4o (0.579) \\
  & \cellcolor{jdfill}Gemini-2.5 & \cellcolor{jdfill}Sem.\ Neg.\ (0.571) & \cellcolor{jdfill}Seq.\ Prob.\ (0.530) & \cellcolor{jdfill}GPT-4o (0.541) \\
\midrule
\multirow{3}{*}{\rotatebox[origin=c]{90}{SelfAw.}}
  & \cellcolor{bbfill}Llama-3-8B & \cellcolor{bbfill}Non-Cont.\ (0.587) & \cellcolor{bbfill}Seq.\ Prob.\ (0.502) & \cellcolor{bbfill}GPT-4o (0.578) \\
  & \cellcolor{wbfill}GPT-4o     & \cellcolor{wbfill}Exact Match (0.499) & \cellcolor{wbfill}Prob.\ Margin (0.483) & \cellcolor{wbfill}Sonnet (0.528) \\
  & \cellcolor{jdfill}Gemini-2.5 & \cellcolor{jdfill}Exact Match (0.499) & \cellcolor{jdfill}P(True) (0.510) & \cellcolor{jdfill}Sonnet (0.528) \\
\midrule
\multirow{3}{*}{\rotatebox[origin=c]{90}{PopQA}}
  & \cellcolor{bbfill}Llama-3-8B & \cellcolor{bbfill}Non-Cont.\ (0.666) & \cellcolor{bbfill}P(True) (0.695) & \cellcolor{bbfill}GPT-4o (0.770) \\
  & \cellcolor{wbfill}GPT-4o     & \cellcolor{wbfill}Cosine Sim.\ (0.774) & \cellcolor{wbfill}Mean Tok.\ (0.802) & \cellcolor{wbfill}Llama (0.744) \\
  & \cellcolor{jdfill}Gemini-2.5 & \cellcolor{jdfill}Cosine Sim.\ (0.787) & \cellcolor{jdfill}P(True) (0.686) & \cellcolor{jdfill}Sonnet (0.728) \\
\bottomrule
\end{tabular}}%
\end{table}

\subsection{Disagreement Analysis and Taxonomy Validation}
\label{sec:disagreement}
We conduct the disagreement analysis across all twelve combinations of model (Llama-3-8B,
GPT-4o, Gemini-2.5-Flash) and dataset (TriviaQA, HaluEval, SelfAware, PopQA).  For each combination, the best-AUROC scorer per
family is selected from all available scorers; Table~\ref{tab:topscorers}
shows the selected representative for each cell (model, dataset, family).
Each sample is then assigned to one of the four taxonomy quadrants by
a Youden's J optimal split on the BB (consistency) and WB (confidence)
axes (\S\ref{sec:setup}, Eq.~\ref{eq:comp_hal}). Throughout this
section, we report the hallucination-restricted complementarity $C_H$
under YJ-threshold split, with per-combination values for
all three family pairs are reported in Table~\ref{tab:ch50}.

\paragraph{Evidential hierarchy: what validates the taxonomy.}
The DECK taxonomy's mechanistic claims are supported at two levels of evidence.
Primary evidence comes from (i)~\textit{judge-involving pairs} (BB--Judge and
WB--Judge), because the Judge score is a third axis not used to define the four
quadrants, making its disagreement distribution a genuinely independent test of the
predicted coverage patterns; and (ii)~\textit{external-signal validation}
(\S\ref{sec:externalsignal}), which checks taxonomy predictions against the
entity-popularity stratification in PopQA without using any scorer output.
Secondary evidence comes from the \textit{BB--WB geometric check}: because the
quadrant cells are defined by the same BB and WB scores used to represent those families,
BB--WB disagreements are constrained by construction to land on the anti-diagonal
(Knotted and Drift). This pattern is therefore a structural inevitability, not a
discovered regularity---it confirms internal consistency but cannot validate the taxonomy
against an independent signal. The quantitative results below are presented in this order:
judge-involving pairs first (primary), BB--WB check second (secondary).

\begin{table}[ht]
\centering
\caption{Hallucination-restricted complementarity $C_H$ under
Youden's J thresholds for all twelve combinations, with $95\%$ bootstrap
CI ($1000$ resamples).
\textit{Bold}\,=\,larger value between (BB, J) and (WB, J).}
\label{tab:ch50}
\renewcommand{\arraystretch}{1.10}
\scriptsize
\setlength{\tabcolsep}{2pt}
\resizebox{\columnwidth}{!}{%
\begin{tabular}{@{}C{0.85cm} L{1.5cm} C{1.4cm} C{1.4cm} C{1.4cm}@{}}
\toprule
\rowcolor{hdrbg}
\textcolor{white}{\textbf{Dataset}} &
\textcolor{white}{\textbf{Model}} &
\textcolor{white}{\textbf{BB, WB}} &
\textcolor{white}{\textbf{BB, J}} &
\textcolor{white}{\textbf{WB, J}} \\
\midrule
\multirow{3}{*}{\rotatebox[origin=c]{90}{TriviaQA}}
  & \cellcolor{bbfill}Llama-3-8B & \cellcolor{bbfill}\onepm{0.249}{0.056} & \cellcolor{bbfill}\onepm{0.229}{0.056} & \cellcolor{bbfill}\onepm{\textbf{0.244}}{0.059} \\
  & \cellcolor{wbfill}GPT-4o & \cellcolor{wbfill}\onepm{0.306}{0.096} & \cellcolor{wbfill}\onepm{\textbf{0.306}}{0.093} & \cellcolor{wbfill}\onepm{0.204}{0.079} \\
  & \cellcolor{jdfill}Gemini-2.5 & \cellcolor{jdfill}\onepm{0.214}{0.085} & \cellcolor{jdfill}\onepm{0.500}{0.108} & \cellcolor{jdfill}\onepm{\textbf{0.571}}{0.104} \\
\midrule
\multirow{3}{*}{\rotatebox[origin=c]{90}{HaluEval}}
  & \cellcolor{bbfill}Llama-3-8B & \cellcolor{bbfill}\onepm{0.328}{0.044} & \cellcolor{bbfill}\onepm{0.282}{0.045} & \cellcolor{bbfill}\onepm{\textbf{0.352}}{0.047} \\
  & \cellcolor{wbfill}GPT-4o & \cellcolor{wbfill}\onepm{0.208}{0.041} & \cellcolor{wbfill}\onepm{\textbf{0.236}}{0.045} & \cellcolor{wbfill}\onepm{0.137}{0.035} \\
  & \cellcolor{jdfill}Gemini-2.5 & \cellcolor{jdfill}\onepm{0.423}{0.052} & \cellcolor{jdfill}\onepm{0.255}{0.047} & \cellcolor{jdfill}\onepm{\textbf{0.408}}{0.052} \\
\midrule
\multirow{3}{*}{\rotatebox[origin=c]{90}{SelfAw.}}
  & \cellcolor{bbfill}Llama-3-8B & \cellcolor{bbfill}\onepm{0.312}{0.040} & \cellcolor{bbfill}\onepm{\textbf{0.315}}{0.043} & \cellcolor{bbfill}\onepm{0.114}{0.029} \\
  & \cellcolor{wbfill}GPT-4o & \cellcolor{wbfill}\onepm{0.291}{0.042} & \cellcolor{wbfill}\onepm{0.044}{0.018} & \cellcolor{wbfill}\onepm{\textbf{0.327}}{0.042} \\
  & \cellcolor{jdfill}Gemini-2.5 & \cellcolor{jdfill}\onepm{0.298}{0.041} & \cellcolor{jdfill}\onepm{0.048}{0.020} & \cellcolor{jdfill}\onepm{\textbf{0.327}}{0.045} \\
\midrule
\multirow{3}{*}{\rotatebox[origin=c]{90}{PopQA}}
  & \cellcolor{bbfill}Llama-3-8B & \cellcolor{bbfill}\onepm{0.347}{0.053} & \cellcolor{bbfill}\onepm{0.292}{0.051} & \cellcolor{bbfill}\onepm{\textbf{0.321}}{0.051} \\
  & \cellcolor{wbfill}GPT-4o & \cellcolor{wbfill}\onepm{0.309}{0.075} & \cellcolor{wbfill}\onepm{\textbf{0.520}}{0.080} & \cellcolor{wbfill}\onepm{0.513}{0.079} \\
  & \cellcolor{jdfill}Gemini-2.5 & \cellcolor{jdfill}\onepm{0.264}{0.063} & \cellcolor{jdfill}\onepm{0.390}{0.075} & \cellcolor{jdfill}\onepm{\textbf{0.412}}{0.073} \\
\bottomrule
\end{tabular}}%
\end{table}

\begin{figure*}[t]
\centering
\includegraphics[width=\textwidth]{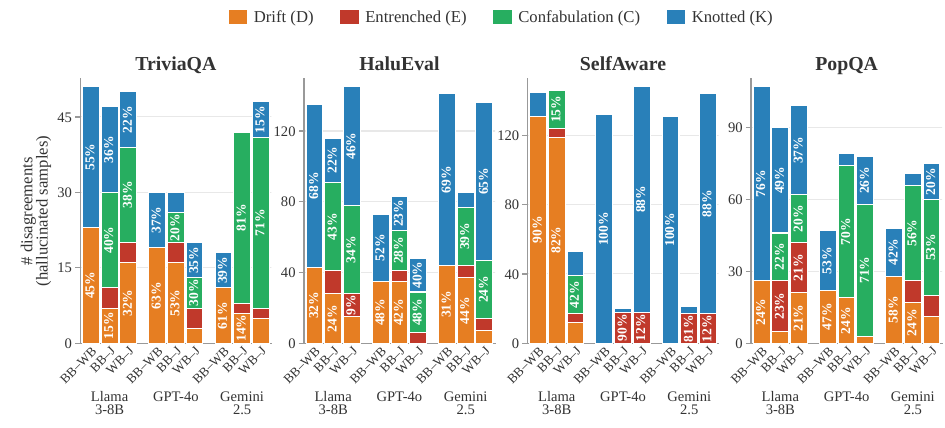}
\caption{Pairwise scorer-family disagreement on hallucinated samples,
one subplot per dataset. Each bar's height is the number of
disagreements for one (model, family pair); segments give the DECK
quadrant share, labelled in \% where space allows (Youden's J split,
matching Table~\ref{tab:ch50}).
}
\label{fig:disagr_2x2}
\end{figure*}

\textit{Judge-involving pairs} (primary validation).
Because the Judge is a third axis not used to define the four
quadrants, its disagreement distribution is the primary test of the
DECK taxonomy. Figure~\ref{fig:disagr_2x2} visualises the per-cell disagreement 
shares across all four datasets. On TriviaQA and HaluEval, judge-involving disagreements
spread across all four quadrants as predicted: 
e.g.\ on Llama-3-8B TriviaQA, BB--Judge disagreements split
between Knotted~(36\%, BB's structural blind spot that the Judge can
recover) and Confabulation~(40\%, where the Judge can diverge from
BB's low-consistency flag); WB--Judge on the same combination splits
between Drift~(32\%, WB's structural blind spot that the Judge can
recover) and Confabulation~(38\%). The predicted BB blind-spot pattern
shows most strongly on Llama-3-8B PopQA, where BB--Judge disagreements
concentrate in K~(49\%) and E~(23\%) --- placing 72\% of the
disagreement set inside BB's predicted high-consistency blind spots.
On GPT-4o/Gemini SelfAware the degenerate two-cell regime constrains
all pairs to Entrenched and Knotted; BB--Judge disagreements nearly
vanish ($n\!\approx\!20$ samples, $\sim\!4\%$ of $N_H$) while WB--Judge
dominates ($n\!\approx\!145$ samples, $\sim\!33\%$ of $N_H$), with BB
and Judge sharing correlated near-random errors and only WB diverging.
Quantitatively, $C_H$ for judge-involving pairs varies substantially
across combinations ($0.044$--$0.571$, see Table~\ref{tab:ch50}) and
tracks the predicted taxonomy profile.

The BB--WB pair is a secondary geometric check rather than independent
validation: BB--WB disagreements are constrained \emph{by construction}
to the anti-diagonal (Knotted and Drift) because the quadrant cells are
defined by the same BB and WB scores. Empirically, this is what we see
(Table~\ref{tab:ch50}, (BB, WB) column; Figure~\ref{fig:disagr_2x2}).

\subsection{External-Signal Validation}
\label{sec:externalsignal}
A second, independent test of the taxonomy uses external labels that
identify the \emph{mechanism} of a hallucination, then ask whether
samples with each mechanism land in the DECK cell that mechanism
predicts. Unlike \S\ref{sec:disagreement}, this test does not depend on
any UQ scorer beyond the (BB, WB) YJ split that defines cell
membership; it asks whether that split is mechanistically informative.

\begin{table*}[!t]
\centering
\small
\renewcommand{\arraystretch}{1.15}
\setlength{\tabcolsep}{4pt}
\caption{External-signal validation: share of samples (\%) in each
DECK cell under a Youden's J split, by external label; Predicted is the
taxonomy's prediction (\S\ref{sec:taxonomy}). $p$: asymptotic $\chi^2$
for SelfAware and HaluEval, permutation test with Cram\'er's $V$ for
PopQA (Appendix~\ref{app:externalsignal}). $^{\S}$\,Minimum expected
count $<5$. $^{\ddagger}$\,Flips to C under cross-fitted thresholds
(Appendix~\ref{app:threshold}).}
\label{tab:external-signal}
\begin{tabular}{@{}lllccccccccc@{}}
\toprule
\textbf{Dataset} & \textbf{Model} & \textbf{External label}
& \textbf{Predicted} & \textbf{Observed}
& $n$ & \textbf{D} (\%) & \textbf{E} (\%) & \textbf{C} (\%) & \textbf{K} (\%)
& \textbf{$p$} & \textbf{$V$} \\
\midrule
SelfAware & Llama-3-8B   & unanswerable    & E   & E             & 500 & 27.0 & \textbf{62.0} &  8.2 &  2.8 & $<$0.001 & -- \\
SelfAware & GPT-4o       & unanswerable    & E   & E             & 500 &  0.0 & \textbf{71.8} &  0.0 & 28.2 & $<$0.001 & -- \\
SelfAware & Gemini-2.5   & unanswerable    & E   & E             & 500 &  0.0 & \textbf{71.2} &  0.0 & 28.8 & $<$0.001 & -- \\
\midrule
HaluEval  & Llama-3-8B   & adversarial     & E/K & \emph{E/C/K}$^{\ddagger}$ & 500 & 10.4 & \textbf{35.8} & \textbf{31.6} & \textbf{22.2} & $<$0.001 & -- \\
HaluEval  & GPT-4o       & adversarial     & E/K & \emph{E}      & 500 & 11.6 & \textbf{64.6} & 14.2 &  9.6 & $<$0.001 & -- \\
HaluEval  & Gemini-2.5   & adversarial     & E/K & E/K           & 500 & 12.4 & \textbf{49.2} &  9.4 & \textbf{29.0} & $<$0.001 & -- \\
\midrule
PopQA     & Llama-3-8B   & popular\_entity & E   & E             &  45 &  2.2 & \textbf{86.7} &  4.4 &  6.7 & $<$0.001$^{\S}$ & 0.43 \\
PopQA     & GPT-4o       & popular\_entity & E   & E             &  45 & 15.6 & \textbf{66.7} &  8.9 &  8.9 & 0.16 & 0.12 \\
PopQA     & Gemini-2.5   & popular\_entity & E   & E             &  45 & 15.6 & \textbf{53.3} & 15.6 & 15.6 & 0.81 & 0.05 \\
\addlinespace[2pt]
PopQA     & Llama-3-8B   & rare\_entity    & C   & \emph{C/K}    & 115 &  8.7 & 13.0 & \textbf{50.4} & \textbf{27.8} & 0.06 & 0.13 \\
PopQA     & GPT-4o       & rare\_entity    & C   & \emph{E/C}    & 115 & 20.0 & \textbf{29.6} & \textbf{38.3} & 12.2 & $<$0.001 & 0.25 \\
PopQA     & Gemini-2.5   & rare\_entity    & C   & \emph{E/C}    & 115 & 11.3 & \textbf{31.3} & \textbf{38.3} & 19.1 & $<$0.001 & 0.20 \\
\bottomrule
\end{tabular}
\end{table*}

\paragraph{Predictions and method.}
Knowledge-gap inputs (SelfAware \emph{unanswerable}) should land in
Entrenched; adversarial inputs (HaluEval) should concentrate in
Knotted or Entrenched; popular-entity inputs (PopQA
\emph{popular\_entity}) in Entrenched; rare-entity inputs (PopQA
\emph{rare\_entity}) in Confabulation. We assign each sample to
a DECK cell using the within-condition Youden's J split, then test
whether the external-label distribution differs from a
naturalistic baseline ($df=3$; full protocol in
Appendix~\ref{app:externalsignal}). PopQA labels use the original
Wikipedia monthly page-view counts $s_{\text{pop}}$ released with
PopQA~\citep{mallen2023popqa}: $s_{\text{pop}}\!\ge\!10000$
$\Rightarrow$~popular ($n{=}45$), $s_{\text{pop}}\!\le\!100$
$\Rightarrow$~rare ($n{=}115$), middle stratum is the naturalistic
baseline ($n{=}339$).

\paragraph{Evidence strength.}
The rows differ sharply in statistical weight. SelfAware and HaluEval
rows use $n{=}500$ per combination and an asymptotic $\chi^2$ test. The
PopQA strata are small --- the popular-entity stratum has only $n{=}45$
(minimum expected count $3.98$ for Llama-3-8B) --- so for all PopQA rows
we replace the asymptotic $\chi^2$ with a Monte-Carlo permutation test
($10^5$ label shuffles) and report Cram\'er's $V$ as effect size. We
treat the \emph{popular-entity} results as \emph{suggestive} and weight
the SelfAware, HaluEval, and rare-entity rows as the primary
external-signal evidence.

\paragraph{Results.}
Table~\ref{tab:external-signal} reports per-cell percentages, $p$-values,
and (for PopQA) effect sizes, with both the taxonomy's predicted
concentration cell and the empirically observed cell(s) shown. Seven
of twelve rows match exactly; the remaining five differ in a single
direction --- four extend the prediction with a secondary cell, and
one (HaluEval $\times$ GPT-4o) is narrower than predicted (E alone
rather than E/K).

SelfAware lands in Entrenched across all three models
($62.0\%$, $71.8\%$, $71.2\%$; all $p<0.001$), matching the predicted
E. Llama-3-8B has a sizable Drift tail ($27.0\%$) alongside E; GPT-4o
and Gemini-2.5-Flash sit in the degenerate two-cell regime of
\S\ref{sec:disagreement}: exact\_match (the best-AUROC BB scorer on
both conditions) returns $0$ on every sample (no response literally
matches an unanswerable question), so the binary BB prediction is
constant. The $\sim\!29\%$ K mass is then an artifact of this constant
BB classifier rather than a separate Knotted observation.

HaluEval adversarial inputs show the most model-dependent pattern.
GPT-4o concentrates sharply in E alone ($64.6\%$, $p<0.001$
--- the table's cleanest single-cell observation, with no detectable
Knotted mass), Gemini-2.5-Flash matches the predicted E/K split
($49.2\%$ E + $29.0\%$ K), and Llama-3-8B spreads across E ($35.8\%$),
C ($31.6\%$), and K ($22.2\%$), suggesting smaller generators mix
Entrenched with Confabulation under adversarial probes. This
Llama-3-8B row is also the one threshold-sensitive row in the table:
its E/C near-tie flips to C under cross-fitted YJ
(Appendix~\ref{app:threshold}).

PopQA \emph{popular-entity} samples (suggestive; $n{=}45$) have
Entrenched as the plurality cell in all three models ($53.3$--$86.7\%$),
matching the predicted E, but the departure from the baseline is
statistically clear only for Llama-3-8B ($86.7\%$, permutation
$p<10^{-4}$, $V\!=\!0.43$). For GPT-4o ($66.7\%$, $p\!=\!0.16$,
$V\!=\!0.12$) and Gemini-2.5-Flash ($53.3\%$, $p\!=\!0.81$,
$V\!=\!0.05$) the effect sizes are small: their middle-stratum
baselines are themselves E-heavy, so the popular subset is not
distinguishable from them at this sample size.
PopQA \emph{rare-entity} samples lead in Confabulation at
$38.3$--$50.4\%$ (Llama-3-8B $50.4\%$, $p\!=\!0.06$, $V\!=\!0.13$;
GPT-4o $38.3\%$, $p<10^{-4}$, $V\!=\!0.25$; Gemini-2.5-Flash $38.3\%$,
$p\!=\!4{\times}10^{-4}$, $V\!=\!0.20$). The secondary cell,
however, diverges across models: GPT-4o and Gemini show Entrenched as
the secondary ($29.6$--$31.3\%$), matching the E/C pattern, whereas
Llama-3-8B has Knotted secondary ($27.8\%$, the C/K pattern). The
GPT-4o/Gemini E secondary has a simple explanation: PopQA questions
use fixed templates (e.g.\ \emph{`What is X's occupation?'}); on rare
entities the model sometimes settles on the most common answer for
that template (e.g.\ \emph{`politician'}) and repeats it confidently
across samples, producing the Entrenched signature. Llama's Knotted
secondary indicates the same template-driven repetition but with
visibly low token-level confidence on the (still-wrong) repeated
answer, consistent with Llama's higher hedging rate.

Together, the four predicted concentrations (SelfAware $\to$ E,
HaluEval $\to$ E/K, PopQA popular $\to$ E, PopQA rare $\to$ C) hold as
the dominant signals across all twelve rows ($11/12$ under
cross-fitted thresholds), with the statistically strongest support
from SelfAware, HaluEval, and the rare-entity stratum. This provides
mechanistic evidence that the YJ-split DECK cells capture genuine
hallucination structure independent of scorer geometry. Model-scale
and content-specific secondary cells emerge as refinements rather than
contradictions.

\subsection{Universal Failure on Knowledge-Gap Hallucinations (SelfAware)}
\label{sec:universalfailure}
\paragraph{Mechanism.}
Output-level UQ scoring is built on three signals --- inter-sample
agreement (BB), token probability (WB), and judge factual review (J)
--- each of which requires a corresponding form of variance in the
generator's behaviour to be informative. When the generator cannot answer an input 
(post-knowledge-cutoff facts, unanswerable questions) but emits a confident, repeatable 
fabrication anyway, every output-level signal collapses at once:
BB sees uniform
agreement, WB sees high token probability, and J shares the same
knowledge gap from common pretraining. Output-level UQ \emph{must} fail
in this regime by construction; the failure is a property of the
(generator, task) pair, not of any specific scorer family.

\paragraph{Empirical confirmation.}
GPT-4o ($90.6\%$), Llama-3-8B ($92.8\%$), and Gemini-2.5-Flash ($88.0\%$)
of SelfAware responses are graded as confident fabrications;
every scorer family then collapses toward chance, and several invert
below it (Table~\ref{tab:auroc5}). On GPT-4o, $13$ of $15$ scorers fall
below $0.5$ and P(True) inverts from $0.675$ on TriviaQA to $0.331$
here, the single worst score in the table, because the self-evaluation
signal mirrors the generator's confident-and-wrong distribution and
anti-correlates with appropriate abstention. We term this regime the
\textbf{universal blind spot of output-level UQ}: \emph{universal} means
across the three output-level families studied here.

\paragraph{Internal-state probe.}
Motivated by evidence that hidden states can encode answerability
\citep{slobodkin2023curious}, we ask a narrower question than that
work: does the output-level failure regime persist in the
activations of our one open-weights generator? This is not a
replication --- model family (Llama-3-8B vs.\ Flan-T5/OPT), metric
(AUROC vs.\ F1), and data (SelfAware vs.\ SQuAD/NQ/MuSiQue) all
differ. We train a logistic probe on Llama-3-8B's last-layer hidden
states, captured under NF4 4-bit quantisation, with TriviaQA
correctness labels. The probe is well calibrated in-distribution
(TriviaQA AUROC $0.72$) but collapses on SelfAware (AUROC $0.44$,
$95\%$ CI $[0.35, 0.52]$); a within-SelfAware $50/50$ probe gives $0.56$,
also CI-overlapping chance. On its own this result is
\emph{inconclusive under the specific conditions tested}: a single
layer, a single model, a single linear probe, and NF4 quantisation,
whose effect on hidden-state fidelity for probing is unstudied. We
therefore replicate it without quantisation (fp16) at layers 8, 16,
24, and 32, keeping inputs and protocol fixed
(Appendix~\ref{app:probe-layers}).

\paragraph{fp16 and multi-layer replication.}
Quantisation does not explain the collapse: fp16 and NF4 last-layer
states are nearly identical (mean cosine similarity $0.98$), and the
fp16 last-layer probe again falls to chance (AUROC $0.43$, CI
$[0.33, 0.52]$). Layer choice matters more. Probes on layers 8 and 24
recover weak but above-chance signal on SelfAware (AUROC $0.63$, CI
$[0.54, 0.72]$; $0.61$, CI $[0.51, 0.71]$), meeting the $0.60$
threshold we fixed in advance, while layer 16 stays at chance ($0.47$).
After Bonferroni correction across the four layers only layer 8
remains distinguishable from chance ($[0.51, 0.73]$). A probe trained
within SelfAware stays at chance at every layer, but with only 18
correct training samples it has little power. We report this as a
boundary of the blind-spot claim: the collapse is robust in the
output-level signals and the final layer, but an intermediate layer
retains weak signal, so the blind spot does not extend to the model's
activations as a whole. Richer internal-state methods --- UQ
heads~\citep{shelmanov2025uqhead}, information-theoretic
estimators~\citep{yadkori2024believe} --- may recover more of it.

\paragraph{Implication.}
The right output-level engineering response is an
abstention envelope \citep{feng2024dont,kalai2025why} that routes
out-of-scope inputs to refusal or retrieval before scoring; the 
ensemble comparison in Appendix~\ref{app:bench}
corroborates this --- no choice of ensemble weights moves
point-estimate AUROC out of $[0.41, 0.61]$ on SelfAware. Where hidden
states are accessible, the intermediate-layer signal suggests
internal-state detectors as a complementary route.

\section{Conclusion}
\label{sec:conclusion}
We introduced the \textbf{DECK taxonomy}, a $2{\times}2$ partition of
LLM hallucinations along inter-sample consistency and token-level
confidence. Unlike prior taxonomies that classify errors by
\emph{what is wrong with the output}, DECK classifies by the
\emph{detectability signature} an error leaves on the (consistency,
confidence) plane. The contribution is not the two axes, which
underpin existing scorers, but the falsifiable per-family blind-spot
map that follows from crossing them (\S\ref{sec:taxonomy}), together
with two independent tests of it.
The disagreement analysis (\S\ref{sec:disagreement}) confirms the
map on judge-involving pairs across nine non-degenerate
(model, dataset) combinations ($C_H$ ranging $0.137$--$0.571$;
Table~\ref{tab:ch50}), and the external-signal validation
(\S\ref{sec:externalsignal}) shows that the four predicted cells hold
as the dominant signals across all twelve
(model, dataset, external-label) rows ($11/12$ under cross-fitted
thresholds, Appendix~\ref{app:threshold}), with the popular-entity
rows suggestive given their small $n$. Model-scale and
content-specific secondary cells emerge as refinements rather than
contradictions. These results are scoped to short-form QA;
\S\ref{sec:longform} sketches how DECK extends to long-form
generation as a claim-level cell mixture.

On knowledge-gap inputs (SelfAware) every output-level scorer
collapses simultaneously: $13$ of $15$ GPT-4o scorers fall below
AUROC\,$0.5$ and P(True) inverts from $0.675$ on TriviaQA to $0.331$,
a \emph{universal blind spot of output-level UQ} that is a property of
the (generator, task) pair, not of any specific family. A linear
probe on Llama-3-8B's last-layer hidden states also falls to chance on
the same regime, with or without 4-bit quantisation (AUROC $0.44$ NF4,
$0.43$ fp16; \S\ref{sec:universalfailure}), but an intermediate layer
retains weak signal (AUROC $0.63$ at layer 8), suggesting that the
blind spot lies in the output-level signals rather than in the model's
activations as a whole. The right engineering
response is an abstention envelope that routes out-of-scope inputs to
refusal or retrieval before scoring rather than a richer ensemble.

\section*{Limitations}
\label{sec:limitations}
Several limitations bound our conclusions.

\textbf{Threshold operationalisation.} The DECK quadrants are defined
by a within-condition Youden's J optimal split on the (BB, WB) plane;
this is base-rate robust but still data-dependent. Appendix~\ref{app:threshold}
quantifies the dependence: under 5-fold cross-fitted YJ the dominant
external-signal cell is unchanged in $11/12$ rows, but quantitative
quadrant shares move by up to $12.4$\,pp, one row
(HaluEval\,/\,Llama-3-8B, an E/C near-tie) is threshold-sensitive,
and F1-optimal thresholds degenerate on high-prevalence datasets.
Calibrated absolute cutoffs and precision-at-$k$ rules remain
untested; future work should examine whether absolute,
calibration-derived thresholds let the taxonomy discriminate domains
more sharply.

\textbf{Judge bias.} While our judge panel mitigates shared-bias
confounds via cross-LLM judges, it does not eliminate them: all judges
share systematic misconceptions from common web-scale pretraining, an
effect the SelfAware results suggest is especially pronounced for
knowledge-gap inputs.

\textbf{Internal-state scope.} Our \emph{universal blind spot} claim
covers the three output-level families. What we can say about
activations rests on linear probes on a single open-weights model
(Llama-3-8B); GPT-4o and Gemini-2.5-Flash do not expose hidden states.
The final-layer collapse replicates without quantisation, but an
intermediate layer retains weak signal (\S\ref{sec:universalfailure},
Appendix~\ref{app:probe-layers}). We tested four layers, only layer 8
survives multiple-comparison correction, and the within-SelfAware
probe is underpowered (36 correct samples in total). Other
internal-state methods --- UQ heads \citep{shelmanov2025uqhead},
information-theoretic estimators \citep{yadkori2024believe},
attention-pattern analyses --- may recover more signal.

\textbf{Short-form QA only.} All four datasets test single-turn,
short-answer factual QA, and every headline claim is scoped to this
setting. In long-form generation a single response mixes claims with
different consistency--confidence profiles; \S\ref{sec:longform}
outlines a claim-level extension, but we do not validate it
empirically.

\textbf{Small strata.} The PopQA popular-entity stratum has $n{=}45$;
even with permutation tests, the GPT-4o and Gemini-2.5-Flash contrasts
are non-significant with small effect sizes ($V\!\le\!0.12$), so the
popular-entity $\to$ Entrenched prediction is supported clearly only
for Llama-3-8B.

\section*{Ethical Considerations}
\paragraph{Potential risks.}
Detector outputs invite over-reliance: a low-uncertainty score is
not a correctness guarantee, and the universal blind spot shows that knowledge-gap inputs
yield confident fabrications that BB, WB, and Judge scorers miss
simultaneously. The DECK cell map is also, in principle, a recipe
for constructing hallucinations that evade specific detector
families.

\section*{Acknowledgements}
We thank the anonymous reviewers and the area chair for feedback
that substantially improved the paper, in particular the threshold-robustness
analysis (Appendix~\ref{app:threshold}), the small-sample statistics of
\S\ref{sec:externalsignal}, the fp16 multi-layer probe replication
(Appendix~\ref{app:probe-layers}), and the long-form extension of
\S\ref{sec:longform}.

\paragraph{Disclosure of LLM Usage.}
The authors used LLMs to assist with refining portions
of the manuscript, and to help write the figure-generation
scripts in the reproducibility package. All experimental design,
results, and final claims are the authors' responsibility.

\bibliography{ref}

\clearpage

\appendix

\section{Scorer Family Reference}
\label{app:scorers}
Table~\ref{tab:scorers} summarizes the three scorer families evaluated in this paper and their
compatibility with each generator model.  Black-box and judge scorers require only sampling
access and can therefore be applied to any of the four generators.  White-box scorers depend
on per-token log-probabilities, which Llama-3-8B (open-weights, run locally), GPT-4o (via
the OpenAI API \texttt{logprobs} parameter), and Gemini-2.5-Flash (via the Google AI
\texttt{logprobs} parameter) all expose.  Claude Sonnet~4.6 does \emph{not}
provide access to token-level log-probabilities through the Anthropic API, so it is used
exclusively as a judge and is excluded from the white-box family.

\begin{table}[ht]
\centering
\caption{Scorer families, logprob requirements, and model applicability.
\checkmark\ = applicable; $\times$ = excluded.}
\label{tab:scorers}
\renewcommand{\arraystretch}{1.3}
\setlength{\tabcolsep}{3pt}
\resizebox{\columnwidth}{!}{%
\begin{tabular}{@{}L{1.7cm} C{1.75cm} C{2.3cm} C{1.3cm} C{1.85cm} C{1.85cm}@{}}
\toprule
\rowcolor{hdrbg}
\textcolor{white}{\textbf{Family}} &
\textcolor{white}{\textbf{Logprobs?}} &
\textcolor{white}{\textbf{Llama-3-8B}} &
\textcolor{white}{\textbf{GPT-4o}} &
\textcolor{white}{\textbf{Gemini-2.5}} &
\textcolor{white}{\textbf{Sonnet-4.6}} \\
\midrule
\rowcolor{bbfill}
\textbf{Black-box} &
\textcolor{stocgreen}{\textbf{No}} &
\textbf{\checkmark} &
\textbf{\checkmark} &
\textbf{\checkmark} &
\textbf{\checkmark} \\
\rowcolor{wbfill}
\textbf{White-box} &
\textcolor{sysred}{\textbf{Yes}} &
\textbf{\checkmark} &
\textbf{\checkmark} &
\textbf{\checkmark} &
\textcolor{sysred}{\textbf{$\times$}} \\
\rowcolor{jdfill}
\textbf{LLM-Judge} &
\textcolor{stocgreen}{\textbf{No}} &
\textbf{\checkmark} &
\textbf{\checkmark} &
\textbf{\checkmark} &
\textbf{\checkmark} \\
\bottomrule
\end{tabular}}
\end{table}

\section{Response Grading Procedures}
\label{app:grading}
Each dataset uses a distinct grading function to assign a binary correctness label
to every model response.  The label is used as the
ground-truth signal for AUROC computation.  All grading is deterministic and
requires no additional model calls.

\subsection*{TriviaQA and HaluEval}

Both datasets supply a reference answer string.  A response is marked correct if
the \emph{normalised} reference appears as a substring of the \emph{normalised}
response.  Normalisation lowercases the text, strips English articles (\emph{a},
\emph{an}, \emph{the}), removes punctuation, and collapses whitespace---the same
procedure used in the original TriviaQA evaluation~\citep{joshi2017triviaqa}.

\begin{tcolorbox}[codebox, title=TriviaQA / HaluEval grading]
\begin{lstlisting}[style=pysnippet]
import re, string

def normalize(text: str) -> str:
    """Lower, strip articles, strip punctuation, collapse spaces."""
    text = text.lower()
    text = re.sub(r'\b(a|an|the)\b', ' ', text)
    text = text.translate(
        str.maketrans('', '', string.punctuation))
    return ' '.join(text.split())

def is_correct(response: str, ground_truth: str) -> bool:
    """True if normalised ground truth is a substring of response."""
    return normalize(ground_truth) in normalize(response)

df['response_correct'] = df.apply(
    lambda row: is_correct(row['response'], row['ground_truths']),
    axis=1)
\end{lstlisting}
\end{tcolorbox}

\medskip\noindent
HaluEval uses the same function because each sample in the
\texttt{pminervini/HaluEval} \texttt{qa\_samples} split provides a single
reference answer field (\texttt{answer}), identical in structure to TriviaQA.

\subsection*{SelfAware}

SelfAware contains questions that are \emph{genuinely unanswerable}.  A response
is graded correct if it appropriately expresses uncertainty or inability to answer,
detected via a set of regular-expression patterns that cover common refusal and
hedging phrasings.

\begin{tcolorbox}[codebox, title=SelfAware grading]
\begin{lstlisting}[style=pysnippet]
UNCERTAINTY_PATTERNS = [
    r"i don'?t know",
    r"i do not know",
    r"i'?m not sure",
    r"i cannot (answer|say|tell|determine|confirm)",
    r"i'?m unable to",
    r"no (definitive|clear|single|known) answer",
    r"(unclear|unknown|uncertain|unanswerable"
     r"|impossible to (say|know|determine))",
    r"there('?s| is) no "
     r"(way|information|record|evidence|known)",
    r"(cannot|can'?t) be "
     r"(determined|confirmed|verified|known)",
    r"(it'?s|this is) "
     r"(unclear|uncertain|unknown|not known|debatable)",
]

def is_correct_unanswerable(response: str) -> bool:
    """True if the response expresses appropriate uncertainty."""
    r = response.lower()
    return any(re.search(p, r) for p in UNCERTAINTY_PATTERNS)

df['response_correct'] = df['response'].apply(
    is_correct_unanswerable)
\end{lstlisting}
\end{tcolorbox}

\medskip\noindent
Only the 500 \emph{unanswerable} questions from
the SelfAware dataset are used; answerable questions are excluded because the focus
of this dataset split is specifically on knowledge-gap hallucinations.

\subsection*{PopQA}

PopQA is an entity-centric open-domain QA dataset~\citep{mallen2023popqa}.  Each sample
includes a \texttt{possible\_answers} field listing all acceptable surface forms of the
correct answer.  A response is marked correct if the \emph{normalised} form of any element
in \texttt{possible\_answers} appears as a substring of the \emph{normalised} response,
using the same normalisation procedure as TriviaQA.

\begin{tcolorbox}[codebox, title=PopQA grading]
\begin{lstlisting}[style=pysnippet]
def is_correct_popqa(response: str,
                     possible_answers: list) -> bool:
    """True if any normalised alias is a substring of response."""
    r_norm = normalize(response)
    return any(normalize(ans) in r_norm
               for ans in possible_answers)

df['response_correct'] = df.apply(
    lambda row: is_correct_popqa(
        row['response'], row['possible_answers']),
    axis=1)
\end{lstlisting}
\end{tcolorbox}

\medskip\noindent
The \texttt{possible\_answers} field may contain multiple aliases (e.g.\
\emph{``United States''}, \emph{``USA''}, \emph{``US''}).  Any matching alias is
sufficient for a correct label.

\subsection*{Dataset summary}

\begin{table}[ht]
\centering
\caption{Grading scheme and accuracy per dataset--model combination.}
\label{tab:grading}
\renewcommand{\arraystretch}{1.2}
\setlength{\tabcolsep}{3pt}
\resizebox{\columnwidth}{!}{%
\begin{tabular}{@{}L{1.8cm} L{2.2cm} C{1.1cm} C{1.1cm} C{1.2cm}@{}}
\toprule
\rowcolor{hdrbg}
\textcolor{white}{\textbf{Dataset}} &
\textcolor{white}{\textbf{Grading method}} &
\textcolor{white}{\textbf{Llama}} &
\textcolor{white}{\textbf{GPT-4o}} &
\textcolor{white}{\textbf{Gemini}} \\
\midrule
\rowcolor{bbfill}
TriviaQA  & Norm.\ substring & 59.0\% & 80.4\% & 83.2\% \\
\rowcolor{wbfill}
HaluEval  & Norm.\ substring & 17.6\% & 29.8\% & 33.4\% \\
\rowcolor{jdfill}
SelfAware & Regex uncertainty & 7.2\%  & 9.4\%  & 12.0\% \\
\rowcolor{bbfill}
PopQA     & Alias substring  & 38.3\% & 69.6\% & 63.5\% \\
\bottomrule
\end{tabular}}
\end{table}

\section{Full AUROC Results --- All Datasets and Models}
\label{app:auroc}
\begin{figure*}[ht]
\centering
\includegraphics[width=\textwidth]{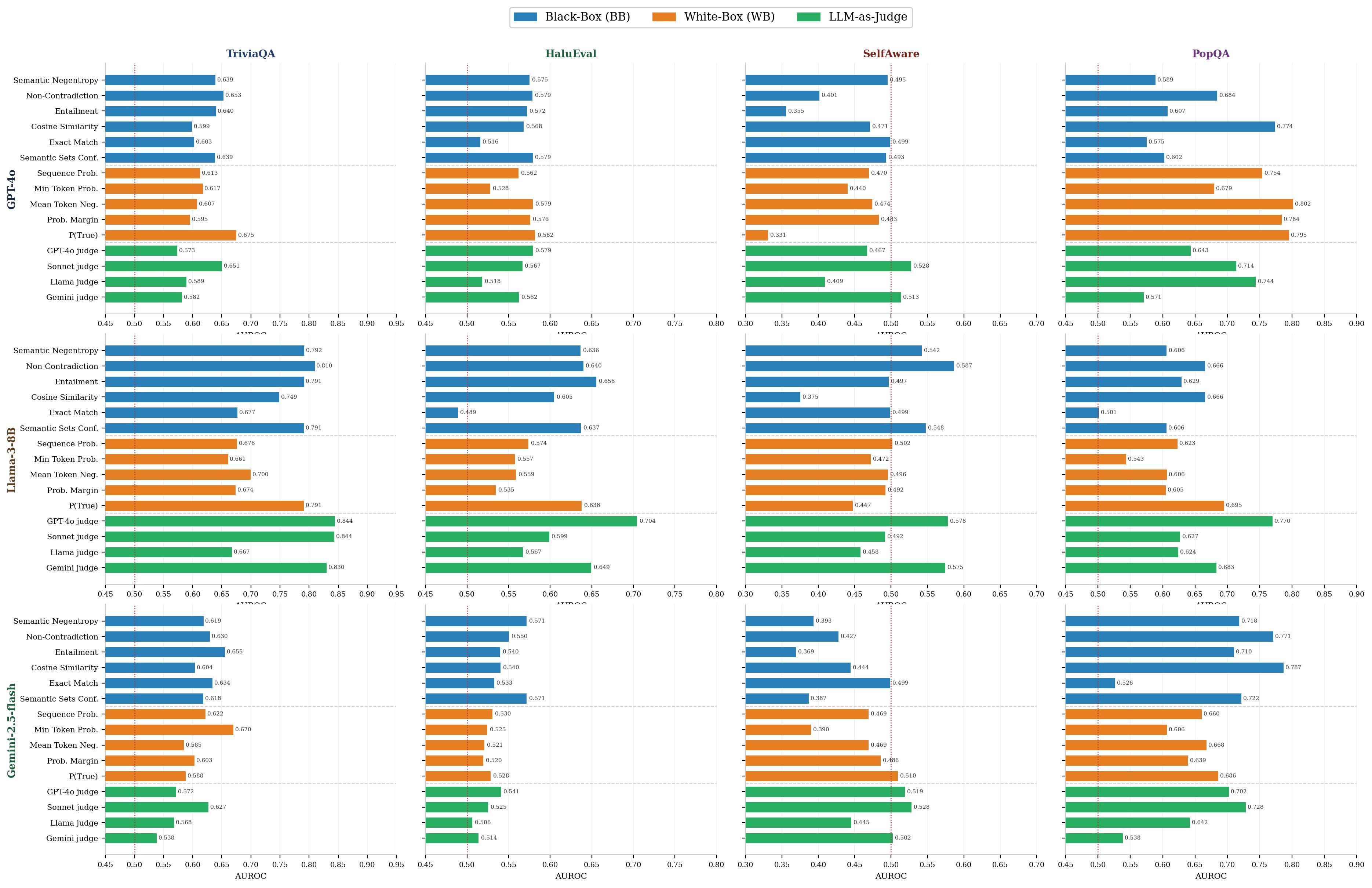}
\caption{Per-method AUROC across all twelve dataset--model combinations
(rows: GPT-4o, Llama-3-8B, Gemini-2.5-Flash; columns: TriviaQA, HaluEval, SelfAware, PopQA).
Methods are grouped by scorer family (blue\,=\,BB, orange\,=\,WB, green\,=\,Judge).
Exact values in Table~\ref{tab:auroc5}.}
\label{fig:auroc_all}
\end{figure*}

Table~\ref{tab:auroc5} reports AUROC for all 15 scorer conditions across all twelve
dataset--model combinations (three models $\times$ four datasets: TriviaQA,
HaluEval, SelfAware, PopQA). The top scorer per family is shown in Figure~\ref{fig:auroc_best}
(in \S\ref{sec:results}); Figure~\ref{fig:auroc_all} visualizes all twelve
dataset--model combinations as horizontal bar charts.
Each cell is mean$\,\pm\,$half-width of a 95\% bootstrap CI ($1000$ resamples, seed 42).

\providecommand{\twoline}[2]{\begin{tabular}[c]{@{}c@{}}#1 \\[-1pt] {\scriptsize $\pm$#2}\end{tabular}}

\begin{table*}[ht]
\centering
\caption{Per-method AUROC with $95\%$ bootstrap CI half-widths across all twelve dataset--model combinations (three models
$\times$ four datasets). Each cell is the AUROC mean (top) with the bootstrap CI half-width below in {\scriptsize\textit{small font}} ($1000$ resamples).
\textbf{Bold}\,=\,best in family per column.
Header shading: {\protect\colorbox{trivhdr}{\textcolor{white}{\strut Triv}}}\,=\,TriviaQA,
{\protect\colorbox{haluhdr}{\textcolor{white}{\strut Halu}}}\,=\,HaluEval,
{\protect\colorbox{selfhdr}{\textcolor{white}{\strut Self}}}\,=\,SelfAware,
{\protect\colorbox{popqahdr}{\textcolor{white}{\strut Pop}}}\,=\,PopQA.
Row shading: {\protect\colorbox{bbfill}{\strut\,}}\,=\,BB,
{\protect\colorbox{wbfill}{\strut\,}}\,=\,WB,
{\protect\colorbox{jdfill}{\strut\,}}\,=\,Judge.}
\label{tab:auroc5}
\renewcommand{\arraystretch}{1.05}
\resizebox{\linewidth}{!}{%
\setlength{\tabcolsep}{3pt}
\begin{tabular}{%
  L{2.4cm}%
  C{1.15cm}%
  C{1.15cm}%
  C{1.15cm}%
  C{1.15cm}%
  C{1.15cm}%
  C{1.15cm}%
  C{1.15cm}%
  C{1.15cm}%
  C{1.15cm}%
  C{1.15cm}%
  C{1.15cm}%
  C{1.15cm}%
}
\toprule
\rowcolor{hdrbg}
\textcolor{white}{\textbf{Scorer}} &
  \cellcolor{trivhdr}\textcolor{white}{\textbf{Llama}\newline\textbf{Triv.}} &
  \cellcolor{trivhdr}\textcolor{white}{\textbf{GPT}\newline\textbf{Triv.}} &
  \cellcolor{trivhdr}\textcolor{white}{\textbf{Gem.}\newline\textbf{Triv.}} &
  \cellcolor{haluhdr}\textcolor{white}{\textbf{Llama}\newline\textbf{Halu.}} &
  \cellcolor{haluhdr}\textcolor{white}{\textbf{GPT}\newline\textbf{Halu.}} &
  \cellcolor{haluhdr}\textcolor{white}{\textbf{Gem.}\newline\textbf{Halu.}} &
  \cellcolor{selfhdr}\textcolor{white}{\textbf{Llama}\newline\textbf{Self.}} &
  \cellcolor{selfhdr}\textcolor{white}{\textbf{GPT}\newline\textbf{Self.}} &
  \cellcolor{selfhdr}\textcolor{white}{\textbf{Gem.}\newline\textbf{Self.}} &
  \cellcolor{popqahdr}\textcolor{white}{\textbf{Llama}\newline\textbf{PopQA}} &
  \cellcolor{popqahdr}\textcolor{white}{\textbf{GPT}\newline\textbf{PopQA}} &
  \cellcolor{popqahdr}\textcolor{white}{\textbf{Gem.}\newline\textbf{PopQA}} \\
\midrule
\rowcolor{bbfill}Non-Cont. & \twoline{\textbf{0.810}}{0.039} & \twoline{\textbf{0.653}}{0.066} & \twoline{0.630}{0.069} & \twoline{0.640}{0.056} & \twoline{0.579}{0.050} & \twoline{0.550}{0.052} & \twoline{\textbf{0.587}}{0.081} & \twoline{0.401}{0.075} & \twoline{0.427}{0.073} & \twoline{\textbf{0.666}}{0.046} & \twoline{0.684}{0.049} & \twoline{0.771}{0.045} \\
\rowcolor{bbfill}Semantic Neg. & \twoline{0.792}{0.040} & \twoline{0.639}{0.054} & \twoline{0.619}{0.052} & \twoline{0.636}{0.062} & \twoline{0.575}{0.050} & \twoline{\textbf{0.571}}{0.045} & \twoline{0.542}{0.086} & \twoline{0.495}{0.077} & \twoline{0.393}{0.078} & \twoline{0.606}{0.051} & \twoline{0.589}{0.055} & \twoline{0.718}{0.043} \\
\rowcolor{bbfill}Entailment & \twoline{0.791}{0.040} & \twoline{0.640}{0.062} & \twoline{\textbf{0.655}}{0.064} & \twoline{\textbf{0.656}}{0.057} & \twoline{0.572}{0.050} & \twoline{0.540}{0.053} & \twoline{0.497}{0.102} & \twoline{0.355}{0.071} & \twoline{0.369}{0.069} & \twoline{0.629}{0.049} & \twoline{0.607}{0.054} & \twoline{0.710}{0.047} \\
\rowcolor{bbfill}Sem.\ Sets Conf. & \twoline{0.791}{0.039} & \twoline{0.639}{0.055} & \twoline{0.618}{0.058} & \twoline{0.637}{0.058} & \twoline{\textbf{0.579}}{0.047} & \twoline{0.571}{0.047} & \twoline{0.548}{0.088} & \twoline{0.493}{0.073} & \twoline{0.387}{0.074} & \twoline{0.606}{0.048} & \twoline{0.602}{0.051} & \twoline{0.722}{0.048} \\
\rowcolor{bbfill}Cosine Sim. & \twoline{0.749}{0.043} & \twoline{0.599}{0.058} & \twoline{0.604}{0.063} & \twoline{0.605}{0.054} & \twoline{0.568}{0.051} & \twoline{0.540}{0.052} & \twoline{0.375}{0.093} & \twoline{0.471}{0.084} & \twoline{0.444}{0.078} & \twoline{0.666}{0.047} & \twoline{\textbf{0.774}}{0.044} & \twoline{\textbf{0.787}}{0.044} \\
\rowcolor{bbfill}Exact Match & \twoline{0.677}{0.041} & \twoline{0.603}{0.059} & \twoline{0.634}{0.064} & \twoline{0.489}{0.014} & \twoline{0.516}{0.045} & \twoline{0.533}{0.043} & \twoline{0.499}{0.002} & \twoline{\textbf{0.499}}{0.002} & \twoline{\textbf{0.499}}{0.002} & \twoline{0.501}{0.011} & \twoline{0.575}{0.027} & \twoline{0.526}{0.028} \\
\midrule
\rowcolor{wbfill}P(True) & \twoline{\textbf{0.791}}{0.040} & \twoline{\textbf{0.675}}{0.066} & \twoline{0.588}{0.070} & \twoline{\textbf{0.638}}{0.059} & \twoline{\textbf{0.582}}{0.051} & \twoline{0.528}{0.051} & \twoline{0.447}{0.085} & \twoline{0.331}{0.065} & \twoline{\textbf{0.510}}{0.075} & \twoline{\textbf{0.695}}{0.048} & \twoline{0.795}{0.045} & \twoline{\textbf{0.686}}{0.051} \\
\rowcolor{wbfill}Mean Token Neg. & \twoline{0.700}{0.044} & \twoline{0.607}{0.060} & \twoline{0.585}{0.061} & \twoline{0.559}{0.066} & \twoline{0.579}{0.051} & \twoline{0.521}{0.055} & \twoline{0.496}{0.102} & \twoline{0.474}{0.079} & \twoline{0.469}{0.083} & \twoline{0.606}{0.052} & \twoline{\textbf{0.802}}{0.040} & \twoline{0.668}{0.051} \\
\rowcolor{wbfill}Sequence Prob. & \twoline{0.676}{0.048} & \twoline{0.613}{0.062} & \twoline{0.622}{0.066} & \twoline{0.574}{0.064} & \twoline{0.562}{0.054} & \twoline{\textbf{0.530}}{0.054} & \twoline{\textbf{0.502}}{0.089} & \twoline{0.470}{0.077} & \twoline{0.469}{0.078} & \twoline{0.623}{0.052} & \twoline{0.754}{0.044} & \twoline{0.660}{0.050} \\
\rowcolor{wbfill}Prob.\ Margin & \twoline{0.674}{0.046} & \twoline{0.595}{0.058} & \twoline{0.603}{0.062} & \twoline{0.535}{0.066} & \twoline{0.576}{0.052} & \twoline{0.520}{0.054} & \twoline{0.492}{0.095} & \twoline{\textbf{0.483}}{0.081} & \twoline{0.486}{0.079} & \twoline{0.605}{0.047} & \twoline{0.784}{0.041} & \twoline{0.639}{0.053} \\
\rowcolor{wbfill}Min Token Prob. & \twoline{0.661}{0.047} & \twoline{0.617}{0.065} & \twoline{\textbf{0.670}}{0.066} & \twoline{0.557}{0.067} & \twoline{0.528}{0.056} & \twoline{0.525}{0.052} & \twoline{0.472}{0.085} & \twoline{0.440}{0.077} & \twoline{0.390}{0.082} & \twoline{0.543}{0.053} & \twoline{0.679}{0.049} & \twoline{0.606}{0.050} \\
\midrule
\rowcolor{jdfill}GPT-4o judge & \twoline{\textbf{0.844}}{0.035} & \twoline{0.573}{0.039} & \twoline{0.572}{0.044} & \twoline{\textbf{0.704}}{0.052} & \twoline{\textbf{0.579}}{0.033} & \twoline{\textbf{0.541}}{0.029} & \twoline{\textbf{0.578}}{0.076} & \twoline{0.467}{0.054} & \twoline{0.519}{0.020} & \twoline{\textbf{0.770}}{0.042} & \twoline{0.643}{0.037} & \twoline{0.702}{0.039} \\
\rowcolor{jdfill}Sonnet judge & \twoline{0.844}{0.032} & \twoline{\textbf{0.651}}{0.053} & \twoline{\textbf{0.627}}{0.052} & \twoline{0.599}{0.058} & \twoline{0.567}{0.047} & \twoline{0.525}{0.051} & \twoline{0.492}{0.085} & \twoline{\textbf{0.528}}{0.056} & \twoline{\textbf{0.528}}{0.034} & \twoline{0.627}{0.048} & \twoline{0.714}{0.045} & \twoline{\textbf{0.728}}{0.046} \\
\rowcolor{jdfill}Llama judge & \twoline{0.667}{0.044} & \twoline{0.589}{0.058} & \twoline{0.568}{0.064} & \twoline{0.567}{0.059} & \twoline{0.518}{0.048} & \twoline{0.506}{0.046} & \twoline{0.458}{0.069} & \twoline{0.409}{0.059} & \twoline{0.445}{0.063} & \twoline{0.624}{0.044} & \twoline{\textbf{0.744}}{0.047} & \twoline{0.642}{0.044} \\
\rowcolor{jdfill}Gemini judge & \twoline{0.830}{0.037} & \twoline{0.582}{0.043} & \twoline{0.538}{0.032} & \twoline{0.649}{0.052} & \twoline{0.562}{0.029} & \twoline{0.514}{0.021} & \twoline{0.575}{0.075} & \twoline{0.513}{0.007} & \twoline{0.502}{0.003} & \twoline{0.683}{0.043} & \twoline{0.571}{0.032} & \twoline{0.538}{0.025} \\
\bottomrule
\end{tabular}}%
\end{table*}

The three SelfAware columns (dark red header) reveal a consistent blind spot across all models.
For GPT-4o, thirteen of fifteen scorers fall below~0.5 (best: Sonnet judge 0.528), with
P(True) producing the single worst score (0.331) in this subtable.  For Llama-3-8B,
the failure is less extreme but still substantial: nine of fifteen scorers fall below~0.5,
with the GPT-4o cross-judge (0.578), Gemini cross-judge (0.575), and Non-Contradiction (0.587) providing modest
above-chance signal.  Gemini-2.5-Flash sits between the two: all scorers cluster
between 0.37 and 0.53, with the judges (GPT-4o 0.519, Sonnet 0.528, Gemini self-judge 0.502) and P(True) (0.510) marginally above
chance while BB scorers mostly fall below.  All three profiles confirm
the universal blind spot of \S\ref{sec:universalfailure}---no scorer family provides reliable
detection when the model is systematically and confidently wrong.

On TriviaQA and HaluEval, Gemini-2.5-Flash AUROC is uniformly compressed relative to
Llama-3-8B and comparable to or slightly below GPT-4o: on TriviaQA the best BB (Entailment,
0.655) and WB (Min Token Prob., 0.670) scorers match GPT-4os top scores, while on HaluEval
all families collapse toward 0.53--0.57, roughly 0.07--0.13 points below the Llama leader.
The Gemini HaluEval compression likely reflects the models higher baseline accuracy (33.4\%
vs.\ 17.6\% for Llama): with fewer errors to detect, all uncertainty signals operate on a
harder residual set of hallucinations.

The three PopQA columns (purple header) stand out as the highest-AUROC region of the table
for GPT-4o and Gemini-2.5-Flash; the ceiling is lower for Llama-3-8B
($0.67$--$0.77$ vs.\ $0.69$--$0.80$ for GPT-4o and Gemini).
Cosine Sim.\ leads the BB family for GPT-4o (0.774) and Gemini (0.787);
for Llama-3-8B, BB peaks at 0.666 (Non-Contradiction and Cosine Sim.\
effectively tied) --- about 0.10 AUROC below GPT-4o and Gemini,
consistent with Llama's higher hedging rate on PopQA.
WB scorers are notably asymmetric across models: GPT-4os Mean Token Neg.\
(0.802) substantially outperforms Llamas P(True) (0.695) and Geminis P(True) (0.686),
consistent with GPT-4os higher PopQA baseline accuracy providing better-calibrated
token-level probabilities.  The Judge family is led by the GPT-4o cross-judge for
Llama-3-8B (0.770), the Llama cross-judge for GPT-4o (0.744), and Sonnet for
Gemini-2.5-Flash (0.728); cross-model judges consistently provide complementary
signal to the generators' own uncertainty on entity-centric queries.

\section{Step-by-Step Computation of the Pairwise Complementarity Score}
\label{app:complementarity}
This appendix walks through the exact computation of the hallucination-restricted
pairwise complementarity score $C_H(A,B)$ as used in
\S\ref{sec:disagreement}, from raw scorer outputs to the per-cell values in
Table~\ref{tab:ch50}.

\subsection*{Step 1: Select the best scorer per family}

For a given (model, dataset) combination (e.g.\ Llama-3-8B on TriviaQA),
each scorer family (BB, WB, Judge) is represented by its single
highest-AUROC method on that split. The representative is chosen from
Table~\ref{tab:topscorers}; this avoids hand-selection bias while
ensuring each family contributes its strongest available signal.

\textit{Example.} For Llama-3-8B on TriviaQA the representatives are
Non-Contradiction (BB, AUROC $0.810$), P(True) (WB, AUROC $0.791$), and
the GPT-4o cross-judge (Judge, AUROC $0.844$).

\subsection*{Step 2: Binarise scorer outputs at the Youden's J threshold}

Each representative produces a continuous confidence score for every
sample. A higher score means the scorer considers the response
\emph{more likely correct}. We binarise by the Youden's J optimal
threshold (\S\ref{sec:setup}):
\[
  \hat{y}_i^{(A)} =
  \begin{cases}
    1 & \text{if } s_i^{(A)} \geq \tau^{(A)} \\
    0 & \text{if } s_i^{(A)} <  \tau^{(A)}
  \end{cases}
\]
($1$ = predicted correct; $0$ = predicted hallucination), where
$\tau^{(A)} = \arg\max_\tau [\mathrm{TPR}(\tau) - \mathrm{FPR}(\tau)]$
fit on the full $N=500$ samples of the condition (positive class
$=$ hallucinated). Unlike a median split, YJ does not enforce a fixed
flag rate; the flag rate is whatever maximises balanced accuracy on
that condition.

\subsection*{Step 3: Identify disagreement samples within $\mathcal{H}$}

Let $\mathcal{H} = \{i : \neg\,\text{correct}_i\}$ denote the
hallucinated-sample subset of size $N_H = |\mathcal{H}|$. A sample
$i \in \mathcal{H}$ is a \emph{disagreement} between families $A$ and
$B$ if their binary predictions differ:
\[
  \mathcal{D}^H(A,B) = \bigl\{ i \in \mathcal{H} : \hat{y}_i^{(A)} \neq \hat{y}_i^{(B)} \bigr\}
\]
Disagreements split into two disjoint cases:
\begin{itemize}
  \item $A$ flags, $B$ misses: $\hat{y}_i^{(A)}=0,\; \hat{y}_i^{(B)}=1$
        (A predicts hallucination, B predicts correct).
  \item $B$ flags, $A$ misses: $\hat{y}_i^{(A)}=1,\; \hat{y}_i^{(B)}=0$.
\end{itemize}
Let $n_{01}^H = |\{i \in \mathcal{H} : \hat{y}^{(A)}{=}0,\hat{y}^{(B)}{=}1\}|$
and $n_{10}^H = |\{i \in \mathcal{H} : \hat{y}^{(A)}{=}1,\hat{y}^{(B)}{=}0\}|$,
so $|\mathcal{D}^H| = n_{01}^H + n_{10}^H$.

\subsection*{Step 4: Apply the complementarity formula}

The hallucination-restricted complementarity score
(Eq.~\ref{eq:comp_hal}) is the fraction of $\mathcal{H}$ on which the
two scorers disagree:
\begin{align}
  C_H(A,B) = \frac{|\mathcal{D}^H(A,B)|}{N_H}
           = \frac{n_{01}^H + n_{10}^H}{N_H}.
\end{align}
This is a per-condition rate, not a conditional probability average:
the denominator is fixed at $N_H$ regardless of how many samples
either scorer flags. Because the YJ threshold is fit on the full $N$
samples (not on $\mathcal{H}$ alone), the per-scorer flag rate
within $\mathcal{H}$ can differ from $50\%$ --- typically scorers
flag a higher fraction of hallucinated samples than of correct ones,
which is exactly why $C_H$ usually differs from the
all-samples disagreement rate $|\mathcal{D}|/N$.

\subsection*{Step 5: Worked example}

\textit{Condition: Llama-3-8B on TriviaQA, pair BB vs.\ WB ($N{=}500$,
$N_H{=}205$).}

\begin{enumerate}
  \item Representatives: Non-Contradiction (BB), P(True) (WB).
  \item Youden's J thresholds: $\tau^{(\text{BB})}=0.754$,
        $\tau^{(\text{WB})}=0.996$. Full-$N$ flag rates: $40.0\%$ (BB),
        $42.2\%$ (WB).
  \item Among the $N_H{=}205$ hallucinated samples, BB flags $142$ and
        WB flags $147$.
  \item Count disagreements within $\mathcal{H}$: $|\mathcal{D}^H|=51$
        ($n_{01}^H=23$ where BB flags but WB misses; $n_{10}^H=28$
        where WB flags but BB misses).
  \item Apply formula:
        \[
          C_H(\text{BB},\text{WB}) = \frac{51}{205} = 0.249.
        \]
\end{enumerate}

\subsection*{Step 6: Why $C_H$ differs from the unconditional rate $|\mathcal{D}|/N$}

The pairwise complementarity restricted to $\mathcal{H}$ is not
equivalent to the all-samples rate. Three details matter:
\begin{enumerate}
  \item The YJ thresholds $\tau^{(A)}$ and $\tau^{(B)}$ are computed
        over \emph{all} $N$ samples (positive class $=$ hallucinated),
        not on $\mathcal{H}$ alone. The binarisation is therefore
        identical to the one used in any all-$N$ analysis.
  \item The denominator is $N_H$ rather than $N$, so $C_H$ is a
        conditional rate.
  \item Scorers typically flag a higher fraction of hallucinated
        samples than of correct ones, so the within-$\mathcal{H}$ flag
        rate exceeds the marginal rate; this concentrates agreement on
        $\mathcal{H}$ and pulls $C_H$ \emph{below} the unconditional
        $|\mathcal{D}|/N$.
\end{enumerate}

\textit{Continuing the worked example.}
On Llama-3-8B TriviaQA the unconditional rate is
$|\mathcal{D}|/N = 115/500 = 0.230$, while
$C_H = 51/205 = 0.249$. The two differ because both BB and WB flag
hallucinated samples at a higher rate than the all-$N$ flag rate
($\sim\!70\%$ vs.\ $\sim\!40\%$), concentrating agreement inside
$\mathcal{H}$ but also leaving a non-trivial residue of asymmetric
disagreements ($23$ vs.\ $28$) that the body uses as the primary
evidential signal.

\subsection*{Step 7: Quadrant decomposition of disagreements}

Each disagreement sample is additionally assigned to one of the four
DECK quadrants using the \emph{same} YJ thresholds on the raw
$(s^{(\text{BB})}, s^{(\text{WB})})$ pair --- i.e.\ the cell membership
shown in Figure~\ref{fig:disagr_2x2} uses the YJ split, not a separate
median split:

\begin{center}
\renewcommand{\arraystretch}{1.2}
\scriptsize
\begin{tabular}{@{}lcc@{}}
\toprule
\textbf{Quadrant} & \textbf{BB score} & \textbf{WB score} \\
\midrule
Entrenched (E)     & $\geq \tau^{(\text{BB})}$ & $\geq \tau^{(\text{WB})}$ \\
Knotted (K)        & $\geq \tau^{(\text{BB})}$ & $<     \tau^{(\text{WB})}$ \\
Drift (D)          & $<     \tau^{(\text{BB})}$ & $\geq \tau^{(\text{WB})}$ \\
Confabulation (C)  & $<     \tau^{(\text{BB})}$ & $<     \tau^{(\text{WB})}$ \\
\bottomrule
\end{tabular}
\end{center}

This is the same cell-membership rule used in
\S\ref{sec:taxonomy}, \S\ref{sec:disagreement}, and
\S\ref{sec:externalsignal}; the figure inherits the quadrant
assignments and only restricts attention to $\mathcal{D}^H(A,B)$ for
each pair. The quadrant decomposition is purely descriptive --- it
does not change the value of $C_H(A,B)$ --- but it reveals
\emph{where} in the hallucination space each family pair disagrees,
which is the central diagnostic of the taxonomy validation.

\section{External-Signal Validation Protocol}
\label{app:externalsignal}
This appendix gives the procedure used to populate
Table~\ref{tab:external-signal}.

\paragraph{Cell assignment.}
Within each (model, dataset) condition, compute the Youden's J optimal
threshold $\tau^{(\text{bb})}$ and $\tau^{(\text{wb})}$ over the full
$N\!=\!500$ samples (positive class = hallucinated). Each sample is
assigned to one of \{D, E, C, K\}.
This is the same YJ-threshold split used in
\S\ref{sec:disagreement}; cell membership is therefore consistent across
the disagreement and external-signal analyses.

\paragraph{Significance test.}
For PopQA, a $2 \times 4$ chi-square contingency table contrasts the target
label's (\texttt{popular\_entity} or \texttt{rare\_entity}) cell distribution
against the \texttt{middle} stratum (the naturalistic baseline). For HaluEval,
the contrast is against the TriviaQA stratum re-binned using the current
condition's YJ thresholds. For single-label datasets (SelfAware), the contrast
is against pooled \texttt{natural} samples from \emph{other} datasets evaluated
on the same model, re-binned using the current condition's YJ thresholds.
No continuity correction is applied (Yates's correction is only defined for
$2 \times 2$ tables); a warning is emitted when any expected cell count is
below 5. Cells whose combined target+baseline count is zero (degenerate
collapses to two-cell distributions under YJ) are dropped before the test;
all p-values reported with df $=3$ unless otherwise noted.

\paragraph{Permutation test for PopQA strata.}
Because the PopQA strata are small, their $p$-values in
Table~\ref{tab:external-signal} come from a Monte-Carlo permutation
test rather than the asymptotic $\chi^2$ distribution. We pool the
target stratum with the \texttt{middle} baseline, shuffle stratum
labels $B\!=\!10^5$ times with the margins fixed (equivalently,
multivariate-hypergeometric draws of the target's cell counts; seed
42), recompute the Pearson $\chi^2$ statistic on each shuffle, and
report $p=(1+\#\{\chi^2_{\text{perm}}\ge\chi^2_{\text{obs}}\})/(B+1)$,
floored at $10^{-5}$. Effect size is Cram\'er's
$V=\sqrt{\chi^2/n_{\text{tot}}}$ for the $2{\times}4$ table. The
permutation and asymptotic $p$-values agree closely except for the
Llama-3-8B popular-entity row (minimum expected count $3.98$), where
the asymptotic $p\!=\!1.6{\times}10^{-15}$ overstates the evidence;
the permutation value is $p<10^{-4}$ ($V\!=\!0.43$). No significance
conclusion changes.

\paragraph{Edge cases.}
(i)~If an external label has fewer than $\sim 50$ samples in a condition, the
chi-square is unreliable; we report rates but flag the p-value. (ii)~The test
assumes the BB and WB scorers used for cell assignment are the same
representatives used in \S\ref{sec:disagreement}; using different scorers
would not be a valid replication. (iii)~The test does not control for
response correctness; an incorrect-only version using
Eq.~\ref{eq:comp_hal}'s subset is a more conservative test of mechanistic
distinctiveness.

\paragraph{PopQA popularity labels.}
We use the original Wikipedia monthly page-view counts ($s_{\text{pop}}$) released
with PopQA~\citep{mallen2023popqa}, applied to the
$499$-question slice used throughout this paper. Following Mallen et al.,
samples with $s_{\text{pop}}\!\ge\!10000$ are labelled \texttt{popular\_entity}
($n{=}45$), $s_{\text{pop}}\!\le\!100$ are \texttt{rare\_entity}
($n{=}115$), and the remaining stratum ($100\!<\!s_{\text{pop}}\!<\!10000$,
$n{=}339$) serves as the naturalistic baseline. Because \texttt{popular\_entity}
sits at the chi-square reliability threshold ($n{=}45$, minimum expected count
$3.98$--$5.86$ across models), the popular-entity contrasts should be read as
suggestive rather than definitive; the \texttt{rare\_entity}
contrasts are well above the threshold ($n{=}115$, minimum expected count
$\ge\!10.9$). All HaluEval rows are treated as \texttt{adversarial} (the dataset
is fully curated to test hallucination); the TriviaQA stratum serves as the
naturalistic baseline, re-binned with each HaluEval condition's YJ thresholds.

Table~\ref{tab:external-signal} (in \S\ref{sec:externalsignal}) reports
the resulting per-cell percentages and chi-square statistics under the
above protocol.

\section{The Effect of Model Scale}
\label{app:modelscale}
We compare hallucination-restricted complementarity $C_H$ across
Llama-3-8B, GPT-4o, and Gemini-2.5-Flash (Table~\ref{tab:ch50}) to test
whether scaling shifts the mix of hallucination types in ways the DECK
taxonomy predicts. The pattern is dataset-dependent rather than monotonic
in model scale, and the differences in quadrant decomposition are at least
as informative as the differences in $C_H$ magnitude.

\textbf{TriviaQA (factual open-domain QA).}
Moving from Llama-3-8B to GPT-4o, $C_H(\text{BB},\text{WB})$ rises from
$0.249$ to $0.306$ ($\Delta\!=\!{+}0.057$) and
$C_H(\text{BB},\text{J})$ rises from $0.229$ to $0.306$
($\Delta\!=\!{+}0.077$), while $C_H(\text{WB},\text{J})$ \emph{decreases}
slightly ($0.244 \to 0.204$, $\Delta\!=\!{-}0.040$). The quadrant
decomposition reveals what scaling actually changes: at Llama-3-8B,
BB--Judge disagreements concentrate in Confabulation ($40\%$) and
Knotted ($36\%$); at GPT-4o, they shift sharply to Drift ($53\%$),
indicating that scaling moves the residual disagreement from
low-consistency, low-confidence errors (Confabulation/Knotted) toward
the high-confidence/low-consistency Drift quadrant where WB still sees
signal but BB does not. This is consistent with the
Entrenched-shift hypothesis under DECK: GPT-4o's hallucinations become
more confidently-asserted but Judge-detectable, a regime where the
Judge--BB pair recovers complementarity that WB--Judge cannot.

\textbf{HaluEval (adversarial hallucination detection).}
The scale effect is in the \emph{opposite} direction:
$C_H(\text{BB},\text{WB})$ decreases ($0.328 \to 0.208$,
$\Delta\!=\!{-}0.120$), $C_H(\text{BB},\text{J})$ decreases
($0.282 \to 0.236$), and $C_H(\text{WB},\text{J})$ drops sharply
($0.352 \to 0.137$). All three pairs converge as model scale grows on
adversarial inputs, consistent with the Judge and the larger generator
sharing more pretraining-driven misconceptions on HaluEval's curated
hallucination probes than they do on naturalistic factual QA. GPT-4o
HaluEval's WB--Judge disagreements still split between Confabulation
($48\%$) and Knotted ($40\%$), so where disagreement does occur it
remains taxonomy-consistent; the magnitude simply shrinks.

\textbf{PopQA (entity-centric).}
PopQA shows a substantial scale-driven complementarity gain:
$C_H(\text{BB},\text{J})$ rises from $0.292$ (Llama-3-8B) to $0.520$
(GPT-4o, $\Delta\!=\!{+}0.228$) and $C_H(\text{WB},\text{J})$ from
$0.321$ to $0.513$ ($\Delta\!=\!{+}0.192$). Both judge-involving pairs
grow by roughly $0.20$ in absolute terms (a $1.6$--$1.8\times$ relative
increase from Llama to GPT-4o). The quadrant decomposition holds the predicted shape:
GPT-4o PopQA's BB--Judge disagreements concentrate in Confabulation
($70\%$) --- exactly the cell where BB consistency cannot help and an
independent Judge can. This is the cleanest scale-driven endorsement of
the taxonomy's complementarity prediction in the dataset.

\textbf{Gemini-2.5-Flash: elevated complementarity, weaker detection.}
Gemini's TriviaQA produces the table's two highest judge-involving values
($C_H(\text{BB},\text{J})\!=\!0.500$,
$C_H(\text{WB},\text{J})\!=\!0.571$), and its HaluEval pushes
$C_H(\text{BB},\text{WB})$ to $0.423$. The quadrant decomposition on
Gemini-TriviaQA puts roughly $80\%$ of BB--Judge and $71\%$ of WB--Judge
disagreements in Confabulation alone, indicating that the three families
are \emph{more independently uncertain} on Gemini's outputs rather than
more complementarily correct --- a pattern consistent with a less
self-consistent generator.

\textbf{Practical implication.}
The PopQA and TriviaQA results confirm dataset-adaptive ensembling: on
naturalistic factual QA at scale, adding a Judge to BB or WB yields
meaningful complementarity gains, and the gain concentrates in the cells
DECK predicts (Drift at scale on TriviaQA; Confabulation throughout on
PopQA). On adversarial QA (HaluEval), Judge--generator agreement
\emph{increases} with scale, narrowing the complementarity envelope and
leaving BB--WB as the more robust pair. This dataset asymmetry is one
reason no single ensembling rule dominates across conditions.

\section{Ensemble AUROC Comparison}
\label{app:bench}
Table~\ref{tab:bench-full} reports AUROCs for four
ensembling strategies on every (model, dataset) condition:
\textit{Best J} (highest-AUROC Judge selected on a $50\%$ training
half, evaluated alone on the held-out test half),
\textit{Best J\,+\,Best WB} and \textit{Best J\,+\,Best BB}, and \textit{Weighted Avg.} (over top three scorer). 
The table is descriptive: no single ensembling
strategy dominates across conditions, and the Judge-only column shows
that a single best Judge is consistently beaten
by ensembling --- especially on PopQA where the gap exceeds
$+0.08$ AUROC.
\begin{table*}[t]
\centering
\scriptsize
\caption{Ensemble AUROC on the held-out 50\% test split.
Each cell is mean$\,\pm\,$half-width of a 95\% bootstrap CI ($1000$
resamples). 
\textit{Bold} = best per row;
$\dagger$ = degenerate regime (universal blind spot of output-level UQ).}
\label{tab:bench-full}
\renewcommand{\arraystretch}{1.05}
\begin{tabular}{cl ccccc}
\toprule
Dataset & Model & Best J & Best J + Best WB & Best J + Best BB & Weighted Avg. & Best \\
\midrule
\multirow{3}{*}{\rotatebox[origin=c]{90}{TriviaQA}}
 & Llama-3-8B   & 0.888\,{\scriptsize $\pm$0.041} & 0.926\,{\scriptsize $\pm$0.036} & 0.921\,{\scriptsize $\pm$0.037} & \textbf{0.939}\,{\scriptsize $\pm$0.032} & WA \\
 & GPT-4o       & 0.585\,{\scriptsize $\pm$0.067} & 0.614\,{\scriptsize $\pm$0.090} & 0.638\,{\scriptsize $\pm$0.087} & \textbf{0.642}\,{\scriptsize $\pm$0.088} & WA \\
 & Gem.-2.5-Fl  & 0.658\,{\scriptsize $\pm$0.074} & \textbf{0.693}\,{\scriptsize $\pm$0.085} & 0.685\,{\scriptsize $\pm$0.089} & 0.678\,{\scriptsize $\pm$0.095} & J+WB \\
\midrule
\multirow{3}{*}{\rotatebox[origin=c]{90}{HaluEval}}
 & Llama-3-8B   & 0.687\,{\scriptsize $\pm$0.074} & 0.661\,{\scriptsize $\pm$0.075} & \textbf{0.699}\,{\scriptsize $\pm$0.074} & 0.683\,{\scriptsize $\pm$0.072} & J+BB \\
 & GPT-4o       & 0.570\,{\scriptsize $\pm$0.048} & 0.542\,{\scriptsize $\pm$0.075} & \textbf{0.581}\,{\scriptsize $\pm$0.075} & 0.562\,{\scriptsize $\pm$0.075} & J+BB \\
 & Gem.-2.5-Fl  & 0.549\,{\scriptsize $\pm$0.043} & 0.524\,{\scriptsize $\pm$0.078} & \textbf{0.581}\,{\scriptsize $\pm$0.068} & 0.568\,{\scriptsize $\pm$0.074} & J+BB \\
\midrule
\multirow{3}{*}{\rotatebox[origin=c]{90}{Self.$^\dagger$}}
 & Llama-3-8B   & 0.590\,{\scriptsize $\pm$0.126} & 0.590\,{\scriptsize $\pm$0.126} & \textbf{0.610}\,{\scriptsize $\pm$0.125} & 0.543\,{\scriptsize $\pm$0.144} & J+BB \\
 & GPT-4o       & \textbf{0.511}\,{\scriptsize $\pm$0.009} & 0.473\,{\scriptsize $\pm$0.102} & 0.509\,{\scriptsize $\pm$0.010} & 0.451\,{\scriptsize $\pm$0.094} & J \\
 & Gem.-2.5-Fl  & \textbf{0.501}\,{\scriptsize $\pm$0.062} & 0.459\,{\scriptsize $\pm$0.105} & 0.499\,{\scriptsize $\pm$0.062} & 0.408\,{\scriptsize $\pm$0.101} & J \\
\midrule
\multirow{3}{*}{\rotatebox[origin=c]{90}{PopQA}}
 & Llama-3-8B   & 0.787\,{\scriptsize $\pm$0.057} & 0.798\,{\scriptsize $\pm$0.057} & 0.797\,{\scriptsize $\pm$0.058} & \textbf{0.799}\,{\scriptsize $\pm$0.054} & WA \\
 & GPT-4o       & 0.789\,{\scriptsize $\pm$0.062} & 0.846\,{\scriptsize $\pm$0.055} & 0.841\,{\scriptsize $\pm$0.056} & \textbf{0.857}\,{\scriptsize $\pm$0.057} & WA \\
 & Gem.-2.5-Fl  & 0.681\,{\scriptsize $\pm$0.061} & 0.709\,{\scriptsize $\pm$0.063} & \textbf{0.753}\,{\scriptsize $\pm$0.063} & 0.751\,{\scriptsize $\pm$0.062} & J+BB \\
\bottomrule
\end{tabular}
\end{table*}

\section{Threshold Robustness of the DECK Assignment}
\label{app:threshold}
\begin{table*}[!t]
\centering
\small
\begin{tabular}{llcccccccc}
\toprule
 & & \multicolumn{4}{c}{Dominant cell (full $N$)} & \multicolumn{3}{c}{$\kappa$ vs.\ YJ} \\
\cmidrule(lr){3-6}\cmidrule(lr){7-9}
Dataset & Model & YJ & Median & F1-opt & CV-YJ & Median & F1-opt & CV-YJ \\
\midrule
TriviaQA & Llama-3-8B & E & E/C & E & E & 0.75 & 0.80 & 0.95 \\
HaluEval & Llama-3-8B & E & E/C & C & C$^{\ddagger}$ & 0.85 & 0.00 & 0.83 \\
SelfAware & Llama-3-8B & E & E/C & C & E & 0.36 & 0.00 & 0.83 \\
PopQA & Llama-3-8B & C & E & C & C & 0.78 & 0.25 & 0.94 \\
\midrule
TriviaQA & GPT-4o & E & E & E & E & 0.47 & 0.71 & 0.73 \\
HaluEval & GPT-4o & E & E & C & E & 0.60 & 0.00 & 0.78 \\
SelfAware & GPT-4o & E & E/K & C & E & 0.56 & 0.00 & 0.96 \\
PopQA & GPT-4o & E & E/C & E & E & 0.66 & 0.99 & 0.91 \\
\midrule
TriviaQA & Gemini-2.5-Flash & E & E/C & E & E & 0.90 & 0.79 & 0.79 \\
HaluEval & Gemini-2.5-Flash & E & E & C & E & 0.53 & 0.00 & 0.81 \\
SelfAware & Gemini-2.5-Flash & E & E & C & E & 0.58 & 0.00 & 0.70 \\
PopQA & Gemini-2.5-Flash & E & E & E & E & 0.71 & 0.72 & 0.89 \\
\bottomrule
\end{tabular}
\caption{Threshold robustness of the DECK assignment. Dominant cell of
each condition (full $N$) under the paper's Youden's J thresholds and
three alternatives (median split, F1-optimal, 5-fold cross-fitted YJ,
seed 42), and Cohen's $\kappa$ between per-sample cell labels and the
YJ assignment. E/C and E/K under Median are exact ties forced by the
$50/50$ margins; F1-opt $\kappa\!=\!0$ rows are the flag-everything
degeneracy at $\gtrsim\!2/3$ prevalence. The PopQA\,/\,Llama-3-8B Median
``flip'' is a $0.2$\,pp margin (E $32.5\%$ vs.\ C $32.3\%$).
$\ddagger$\,E/C near-tie discussed in the text.}
\label{tab:threshold-robustness}
\end{table*}

\begin{figure*}[!t]
\centering
\includegraphics[width=\textwidth]{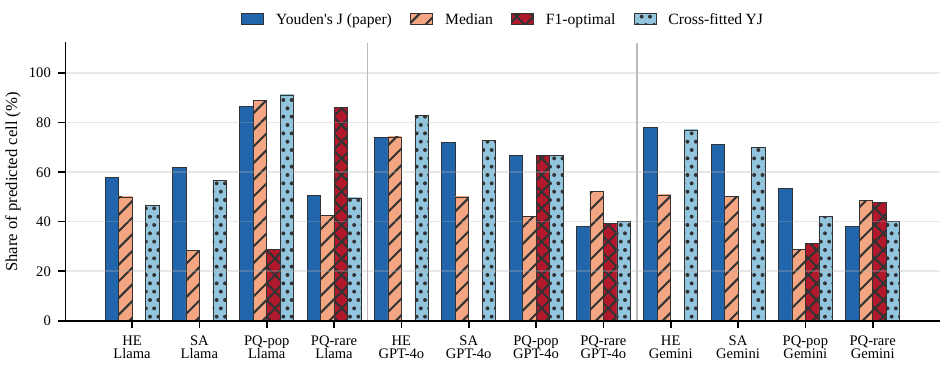}
\caption{Share of each Table~\ref{tab:external-signal} row's
\emph{predicted} DECK cell under four thresholding rules
(SA\,=\,SelfAware unanswerable, predicted E; HE\,=\,HaluEval
adversarial, E+K combined; PQ-pop\,/\,PQ-rare\,=\,PopQA popular\,/\,rare
entity, E\,/\,C). Cross-fitted
YJ stays within $12$\,pp of the paper's YJ share on every row; F1-optimal
collapses to zero on the high-prevalence HaluEval and SelfAware rows
(flag-everything degeneracy).}
\label{fig:threshold-sensitivity}
\end{figure*}

\begin{table*}[t]
\centering
\small
\setlength{\tabcolsep}{8pt}
\begin{tabular}{@{}lcccc@{}}
\toprule
 & \multicolumn{2}{c}{A: TriviaQA$\to$SelfAware} & A2: TriviaQA & B: SelfAware \\
\cmidrule(lr){2-3}
Capture & AUROC [95\% CI] & Bonf.\ CI & AUROC [95\% CI] & AUROC [95\% CI] \\
\midrule
NF4, L32 & 0.44 [0.35, 0.52] & -- & 0.72 [0.66, 0.79] & 0.56 [0.44, 0.68] \\
\midrule
fp16, L8  & \textbf{0.63} [0.54, 0.72] & [0.51, 0.73] & 0.71 [0.64, 0.77] & 0.49 [0.36, 0.62] \\
fp16, L16 & 0.47 [0.38, 0.57] & [0.36, 0.59] & 0.71 [0.64, 0.77] & 0.45 [0.32, 0.57] \\
fp16, L24 & 0.61 [0.51, 0.71] & [0.48, 0.73] & 0.70 [0.64, 0.77] & 0.55 [0.42, 0.67] \\
fp16, L32 & 0.43 [0.33, 0.52] & [0.31, 0.54] & 0.74 [0.68, 0.80] & 0.57 [0.45, 0.69] \\
\bottomrule
\end{tabular}
\caption{Internal-state probe on Llama-3-8B hidden states: the NF4
last-layer probe of \S\ref{sec:universalfailure} and its fp16
multi-layer replication. Bonf.\ CI: protocol-A interval at family-wise
$95\%$ across the four fp16 layers. \textbf{Bold}: the only interval
that excludes $0.5$ after correction. Protocol B has 18 correct
SelfAware samples in each half, so its intervals are wide.}
\label{tab:probe-layers}
\end{table*}

DECK cells are defined by a Youden's J (YJ) split on each scorer axis,
fit on the full $N\!=\!500$ samples of a condition (\S\ref{sec:setup}).
Because this threshold is data-dependent, we re-derive every cell
assignment under three alternative rules and compare each to the
paper's assignment: (i)~a \emph{median split} on each axis;
(ii)~\emph{F1-optimal} thresholds (positive class $=$ hallucinated;
ties broken by the smallest $\tau$); and (iii)~\emph{5-fold
cross-fitted YJ}, in which YJ thresholds are fit on four folds and
applied to the held-out fold, so every sample is assigned
out-of-fold (stratified on correctness, seed 42). The scorer
representatives are unchanged (Table~\ref{tab:topscorers}); only the
thresholds move. Agreement is measured by the dominant cell of each
Table~\ref{tab:external-signal} row and by Cohen's $\kappa$ between
per-sample cell labels (Table~\ref{tab:threshold-robustness},
Figure~\ref{fig:threshold-sensitivity}).

\paragraph{Cross-fitted YJ (the direct test of data-dependent
thresholds).} The dominant cell of $11/12$
Table~\ref{tab:external-signal} rows is unchanged; per-sample
agreement is $\kappa\!=\!0.70$--$0.96$ (mean $0.84$), and the largest
per-cell share shift across the twelve conditions is $12.4$\,pp.
The one exception is HaluEval\,/\,Llama-3-8B, an E/C near-tie
($35.8\%$ vs.\ $31.6\%$) that becomes $30.6\%$ vs.\ $38.4\%$
out-of-fold, because the BB (entailment) axis has two near-tied YJ
optima ($\tau\!\approx\!0.11$ and $\approx\!0.31$) and two of five
folds select the higher one. Table~\ref{tab:external-signal} already
reports this row as mixed (E/C/K); we mark it as threshold-sensitive
($\ddagger$). Strictly, the count of rows whose dominant cell lies in
the predicted set therefore reads $11/12$ under cross-fitted YJ.

\paragraph{Median split.} The dominant cell is unchanged in $9/12$
rows ($\kappa\!=\!0.36$--$0.90$, mean $0.65$). All three exceptions
are exact $50/50$ ties that still contain the YJ cell: when both
axes split exactly $250/250$, the $2{\times}2$ margins force
$\text{E}\!=\!\text{C}$ and $\text{D}\!=\!\text{K}$ identically, so
these ties are an artefact of the rule rather than evidence about
the taxonomy. The largest shift ($33.6$\,pp, SelfAware\,/\,Llama-3-8B)
arises because a fixed $50\%$ flag rate cannot track the condition's
$93\%$ hallucination prevalence.

\paragraph{F1-optimal thresholds.} The dominant cell is unchanged in
only $4/12$ rows (mean $\kappa\!=\!0.35$), but this reflects a known
pathology of F1 under class imbalance rather than instability of the
taxonomy. With hallucination prevalence $\pi$, flagging every sample
achieves $F_1\!=\!2\pi/(1\!+\!\pi)$; on all six HaluEval and SelfAware
conditions ($\pi\!\gtrsim\!2/3$) no interior threshold beats it on either
axis, so every sample is double-flagged, all mass moves to
Confabulation, and $\kappa\!=\!0$. This is why the paper uses the
prevalence-robust YJ criterion. On the six TriviaQA and PopQA
conditions, where F1 does not degenerate, agreement is
$\kappa\!=\!0.25$--$0.99$.

\paragraph{Summary.} Under the two non-degenerate alternatives no
Table~\ref{tab:external-signal} conclusion changes except the
HaluEval\,/\,Llama-3-8B near-tie; quantitative cell shares move, as
anticipated in the Limitations, but the predicted concentrations of
\S\ref{sec:externalsignal} are preserved.

\section{Internal-State Probe: fp16 and Multi-Layer Replication}
\label{app:probe-layers}
We re-captured last-response-token hidden states for the same 500
SelfAware and 500 TriviaQA (prompt, response) pairs used in
\S\ref{sec:universalfailure}, with the same Llama-3-8B-Instruct weights
and chat formatting, but in fp16 without quantisation and at transformer
layers 8, 16, 24, and 32 (of 32). No input exceeded the context limit,
so both captures see identical token sequences. For each layer we fit
the same probe as in \S\ref{sec:universalfailure} (standardised
features, $L_2$-regularised logistic regression with $C\!=\!1$ and
balanced class weights) under the same three protocols:
\textbf{A} trains on TriviaQA and evaluates on SelfAware;
\textbf{A2} evaluates on a held-out $50\%$ TriviaQA split;
\textbf{B} uses a stratified $50/50$ split within SelfAware (seed 42).
CIs are $95\%$ percentile intervals from $1000$ stratified bootstrap
resamples. For protocol~A we also report a Bonferroni-adjusted interval
across the four layers ($\alpha\!=\!0.0125$). NF4 and fp16 last-layer
states have mean cosine similarity $0.98$ on both datasets (5th
percentile $0.97$), so quantisation perturbs the activations only
slightly, and the fp16 layer-32 probe reproduces the NF4 result
(Table~\ref{tab:probe-layers}).

The in-distribution probe (A2) is stable across layers
($0.70$--$0.74$), so every layer encodes correctness on TriviaQA; what
varies is transfer to SelfAware. Transfer fails at layers 16 and 32 and
is weakly positive at layers 8 and 24, and the pattern is not monotonic
in depth. Only layer 8 survives the multiple-comparison correction, and
its lower bound ($0.51$) is close to chance, so we treat the
intermediate-layer signal as weak evidence that the blind spot does not
extend to all activations, not as a usable detector.

\end{document}